\documentclass{article}

\PassOptionsToPackage{numbers, compress}{natbib}
\usepackage[final]{neurips_2023}

\usepackage[utf8]{inputenc} % allow utf-8 input
\usepackage[T1]{fontenc}    % use 8-bit T1 fonts
\usepackage{hyperref}       % hyperlinks
\usepackage{url}            % simple URL typesetting
\usepackage{booktabs}       % professional-quality tables
\usepackage{amsfonts}       % blackboard math symbols
\usepackage{nicefrac}       % compact symbols for 1/2, etc.
\usepackage{microtype}      % microtypography
\usepackage{xcolor}         % colors
\usepackage{qiangstyle}
\usepackage{tcolorbox}
\usepackage{multirow}
\usepackage{booktabs}
\usepackage{MnSymbol}
\usepackage{physics}

\usepackage{mathtools}
\newcommand{\defeq}{\vcentcolon=}

\definecolor{contextcolor}{RGB}{31,160,171}

\usepackage[font=small,labelfont=bf,tableposition=top]{caption}
\DeclareCaptionLabelFormat{andtable}{#1~#2  \&  \tablename~\thetable}

\newcommand{\fs}[1]{\footnotesize $\pm$#1}
\newcommand{\mr}{{\textbf{MR}}}
\newcommand{\dm}{{\bm{$\Delta m\%$}}}

\newcommand{\best}[1]{{\textbf{\textcolor{contextcolor}{#1}}}}
\newcommand{\analysis}{{\textbf{\textcolor{contextcolor}{Findings:}}}}
\newcommand{\evaluations}{{\textbf{\textcolor{contextcolor}{Evaluations:}}}}

\newcommand{\stl}{\textsc{STL}}
\newcommand{\ls}{\textsc{LS}}
\newcommand{\si}{\textsc{SI}}
\newcommand{\rlw}{\textsc{RLW}}
\newcommand{\dwa}{\textsc{DWA}}
\newcommand{\uw}{\textsc{UW}}

\newcommand{\famo}{\textsc{FAMO}}
\newcommand{\graddrop}{\textsc{GradDrop}}

\newcommand{\mgda}{\textsc{MGDA}}
\newcommand{\pcgrad}{\textsc{PCGrad}}
\newcommand{\imtlg}{\textsc{IMTL-G}}
\newcommand{\cagrad}{\textsc{CAGrad}}
\newcommand{\nashmtl}{\textsc{NashMTL}}
\definecolor{myred}{rgb}{0.858, 0.1, 0.1}
\definecolor{myyellow}{rgb}{0.858, 0.6, 0.1}
\definecolor{mydarkred}{rgb}{0.69,0.0,0.1098}

\definecolor{commentcolor}{RGB}{110,154,155}   % define comment color
\newcommand{\PyComment}[1]{\ttfamily\textcolor{commentcolor}{\# #1}}  % add a "#" before the input text "#1"
\newcommand{\PyCode}[1]{\ttfamily\textcolor{black}{#1}} % \ttfamily is the code font

\title{FAMO: Fast Adaptive Multitask Optimization}

\author{%
  $^\dagger$Bo Liu, $^\ddagger$Yihao Feng, $^{\dagger,\mathsection}$Peter Stone, $^\dagger$Qiang Liu\\
  $^\dagger$The University of Texas at Austin, ~$^\ddagger$Salesforce AI Research, $^{\mathsection}$Sony AI\\
  \texttt{\{bliu, pstone, lqiang\}@cs.utexas.edu},~ yihaof@salesforce.com \\
}

\begin{document}

\maketitle

\begin{abstract}
One of the grand enduring goals of AI is to create generalist agents that can learn multiple different tasks from diverse data via multitask learning (MTL). However, in practice, applying gradient descent (GD) on the average loss across all tasks may yield poor multitask performance due to severe under-optimization of certain tasks. Previous approaches that manipulate task gradients for a more balanced loss decrease require storing and computing all task gradients ($\mathcal{O}(k)$ space and time where $k$ is the number of tasks), limiting their use in large-scale scenarios.
In this work, we introduce Fast Adaptive Multitask Optimization (\famo{}), a dynamic weighting method that decreases task losses in a balanced way using $\mathcal{O}(1)$ space and time.
We conduct an extensive set of experiments covering multi-task supervised and reinforcement learning problems.
Our results indicate that \famo{} achieves comparable or superior performance to state-of-the-art gradient manipulation techniques while offering significant improvements in space and computational efficiency. Code is available at \url{https://github.com/Cranial-XIX/FAMO}.
\end{abstract}
\section{Introduction}
\label{sec::intro}
Large models trained on diverse data have advanced both computer vision~\cite{kirillov2023segment} and natural language processing~\cite{bubeck2023sparks}, paving the way for generalist agents capable of multitask learning (MTL)~\cite{caruana1997multitask}. Given the substantial size of these models, it is crucial to design MTL methods that are \emph{effective} in terms of task performance and \emph{efficient} in terms of space and time complexities for managing training costs and environmental impacts. This work explores such methods through the lens of optimization. 

Perhaps the most intuitive way of solving an MTL problem is to optimize the average loss across all tasks. However, in practice, doing so can lead to models with poor multitask performance: a subset of tasks are \emph{severely under-optimized}. A major reason behind such optimization failure is that a subset of tasks are under-optimized because the average gradient constantly results in small (or even negative) progress on these tasks (see details in Section~\ref{sec:optimization-challenge}).

To mitigate this problem, gradient manipulation methods~\cite{yu2020gradient,liu2020towards,chen2020just,liu2021conflict} compute a new update vector in place of the gradient to the average loss, such that all task losses decrease in a more balanced way. The new update vector is often determined by solving an additional optimization problem that involves all task gradients. 
While these approaches exhibit improved performance, they become computationally expensive when the number of tasks and the model size are large~\cite{xin2022current}. 
This is because they require computing and storing all task gradients at each iteration, thus demanding $\mathcal{O}(k)$ space and time complexities, not to mention the overhead introduced by solving the additional optimization problem.
In contrast, the average gradient can be efficiently computed in $\mathcal{O}(1)$ space and time per iteration because one can first average the task losses and then take the gradient of the average loss.\footnote{Here, we refer to the situation where a single data $x$ can be used to compute all task losses.} To this end, we ask the following question:
\begin{center}
$(Q)$~~\emph{Is it possible to design a multi-task learning optimizer that ensures a balanced reduction in losses across all tasks while utilizing $\mathcal{O}(1)$ space and time per iteration?}
\end{center}

\begin{figure}[t!]
    \centering
    \includegraphics[width=\textwidth]{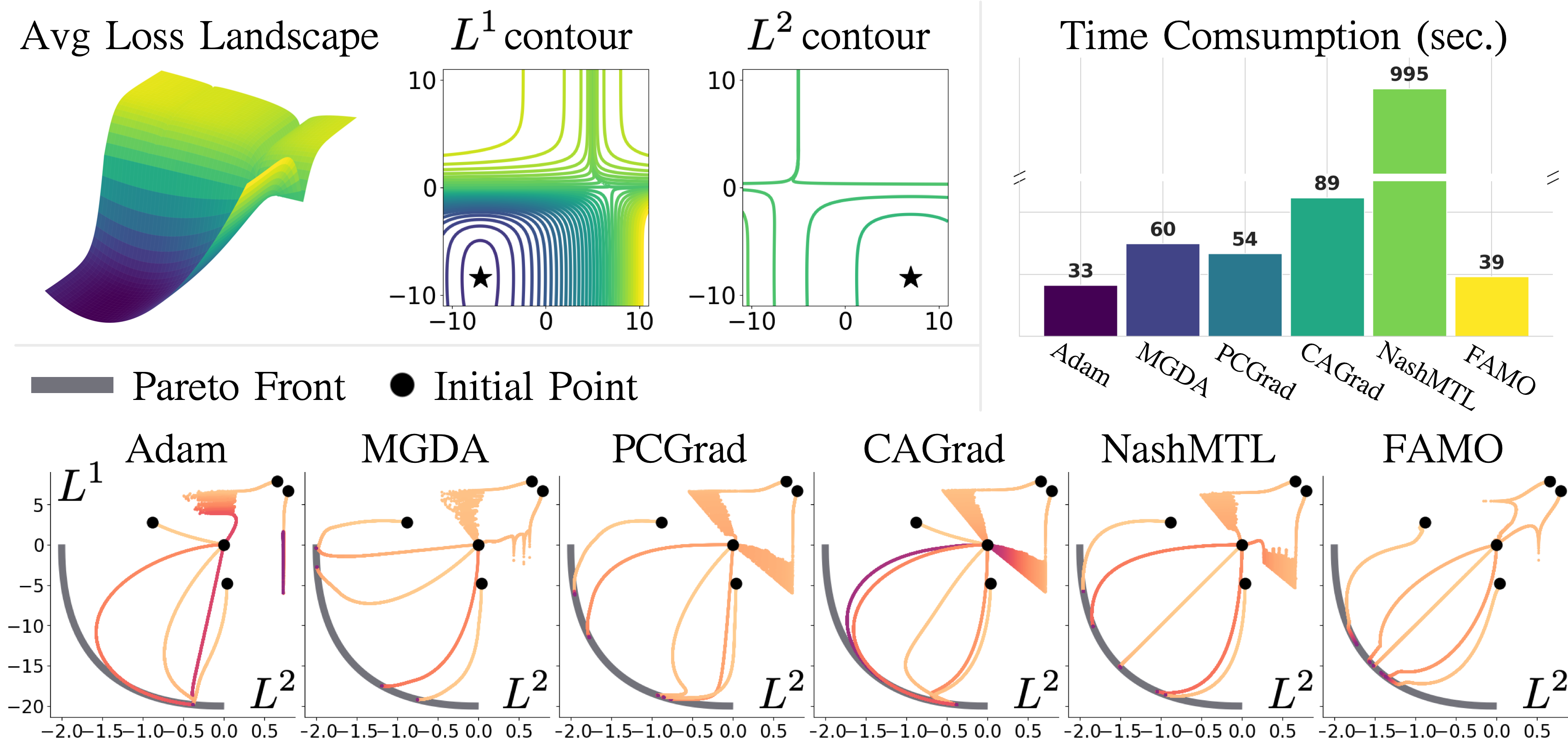}

    \caption{\textbf{Top left:} The loss landscape, and individual task losses of a toy 2-task learning problem ($\filledstar$ represents the minimum of task losses). \textbf{Top right:} the runtime of different MTL methods for 50000 steps. \textbf{Bottom:} the loss trajectories of different MTL methods. \textsc{Adam} fails in 1 out of 5 runs to reach the Pareto front due to CG. \famo{} decreases task losses in a balanced way and is the only method matching the $\mathcal{O}(1)$ space/time complexity of \textsc{Adam}. Experimental details and analysis are provided in Section~\ref{sec::exp-toy}.
    }
    \label{fig:toy}
\end{figure}

In this work, we present Fast Adaptive Multitask Optimization (\famo{}), a simple yet effective adaptive task weighting method to address the above question. On the one hand, \famo{} is designed to ensure that all tasks are optimized with approximately similar progress. On the other hand, \famo{} leverages  the loss history to update the task weighting, hence bypassing the necessity of computing all task gradients. To summarize, our contributions are:
\begin{enumerate}
    \item We introduce \famo{}, an MTL optimizer that decreases task losses approximately at \emph{equal rates} while using only $\mathcal{O}(1)$ space and time per iteration.
    \item We demonstrate that \famo{} performs comparably to or better than existing gradient manipulation methods on a wide range of standard MTL benchmarks, in terms of standard MTL metrics, while being significantly computationally cheaper.
\end{enumerate}
\section{Background}
\label{sec:background}
In this section, we provide the formal definition of multitask learning, then discuss its optimization challenge, and provide a brief overview of the gradient manipulation methods.

\paragraph{Multitask Learning (MTL)} MTL considers optimizing a \emph{single} model with parameter $\x \in \RR^m$ that can perform $k \geq 2$ tasks well, where each task is associated with a loss function $\ell_i(\x): \RR^m \rightarrow \RR_{\geq 0}$.\footnote{In this work, we assume $\forall~i, ~\ell_i(\x) \geq 0$, which is true for typical loss functions including mean square and cross-entropy losses. Note that one can always transform $\ell_i$ to be non-negative if a loss lower bound is known.} Then, it is common to optimize the average loss across all tasks:
\begin{equation}
\min_{\x \in \RR^m} \left \{ \ell_0(\x) \defeq \frac{1}{k}\sum_{i=1}^k \ell_i(\x) \right\}.
\label{eq:mtl-obj}
\end{equation}
\paragraph{Optimization Challenge}
\label{sec:optimization-challenge}
Directly optimizing \eqref{eq:mtl-obj} can result in severe under-optimization of a subset of tasks. A major reason behind this optimization challenge is the ``generalized" conflicting gradient phenomenon, which we explain in the following. At any time step $t$, assume one updates the model parameter using a gradient descent style iterative update: $\x_{t+1} = \x_t - \alpha d_t$ where $\alpha$ is the step size and $d_t$ is the update at time $t$. Then, we say that conflicting gradients (CG)~\citep{liu2021conflict,yu2020gradient} happens if
$$
\exists i,~~ \ell_i(\x_{t+1}) - \ell_i(\x_t) \approx -\alpha \nabla \ell_i(\x_t)^\top d_t > 0.
$$
In other words, certain task's loss is increasing. CG often occurs during optimization and is not inherently detrimental. However, it becomes undesirable when a subset of tasks persistently undergoes under-optimization due to CG. In a more general sense, it is not desirable if a subset of tasks has much slower learning progress compared to the rest of the tasks (even if all task losses are decreasing). This very phenomenon, which we call the ``generalized" conflicting gradient, has spurred previous research to mitigate it at each optimization stage~\cite{yu2020gradient}.

\paragraph{Gradient Manipulation Methods} Gradient manipulation methods aim to decrease all task losses in a more balanced way by finding a new update $d_t$ at each step. $d_t$ is usually a convex combination of task gradients, and therefore the name gradient manipulation (denote $\nabla \ell_{i,t} = \nabla_\x \ell_i(\x_t)$ for short):
\begin{equation}
    d_t = \begin{bmatrix}
    \nabla \ell_{1,t}^\top \\
    \vdots \\
    \nabla \ell_{k,t}^\top 
    \end{bmatrix}^\top w_t,~~~\text{where}~~w_t = \begin{bmatrix}
    w_{1,t} \\
    \vdots \\
    w_{k,t}
    \end{bmatrix} = f\big(\nabla \ell_{1,t}, \dots, \nabla \ell_{k,t}\big) \in \S_k.
    \label{eq:gradmanip}
\end{equation}
Here, $\S_k = \{w \in \RR^k_{\geq 0} \mid w^\top \bm{1} = 1 \}$ is the probabilistic simplex, and $\bm{w}_t$ is the task weighting across all tasks. Please refer to Appendix~\ref{sec::apx-grad-manip} for details of five state-of-the-art gradient manipulation methods (\mgda{}, \pcgrad{}, \cagrad{}, \imtlg{}, \nashmtl{}) and their corresponding $f$.
Note that existing gradient manipulation methods require computing and storing $k$ task gradients before applying $f$ to compute $d_t$, which often involves solving an additional optimization problem. As a result, we say these methods require at least $\mathcal{O}(k)$ space and time complexity, which makes them slow and memory inefficient when $k$ and model size $m$ are large.

\section{Fast Adaptive Multitask Optimization (\famo{})}
\label{sec::method}
In this section, we introduce \famo{} that addresses question~$Q$, which involves two main ideas:
\begin{enumerate}
    \item At each step, decrease all task losses at \emph{an equal rate} as much as possible (Section~\ref{sec::balanced-rate}).
    \item Amortize the computation in 1. over time (Section~\ref{sec:fast-approx}).
\end{enumerate}

\begin{algorithm*}[t!]
    \caption{Fast Adaptive Multitask Optimization (\famo{})}
    \begin{algorithmic}[1]
        \STATE \textbf{Input}: Initial parameter $\theta_0$, task losses $\{\ell_i\}_{i=1}^k$ (ensure that $\ell_i \geq \epsilon > 0$, for instance, by $\ell_i \leftarrow \ell_i - \ell_i^* + \epsilon$, $\ell_i^* = \inf_\x \ell_i(\x)$), learning rate $\alpha$ and $\beta$, and decay $\gamma$ ($=0.001$ by default).
        \STATE ${\xi}_1 \leftarrow {0}$.~~~~~~~~~\best{// initialize the task logits to all zeros}
        \FOR{$t = 1 : T$}
            \STATE Compute ${z}_t = \textbf{Softmax}({\xi}_t)$, e.g., 
            $$
            z_{i,t} = \frac{\exp(\xi_{i,t})}{\sum_{i=1}^k \exp(\xi_{i,t})}.
            $$
            \STATE Update the model parameters:
            $$\theta_{t+1} = \theta_t - \alpha \sum_{i=1}^k \big(c_t\frac{z_{i,t}}{\ell_{i,t}}\big) \nabla \ell_{i,t},~~ \text{where}~~c_t = \big(\sum_{i=1}^k \frac{z_{i,t}}{\ell_{i,t}}\big)^{-1}.$$
            \STATE Update the logits for task weighting:
            \vspace{-5pt}
            $${\xi}_{t+1} = {\xi}_t - \beta \big( {\delta}_t + \gamma {\xi}_t\big)~~\text{where}~~{\delta}_t = 
            \begin{bmatrix}
                \nabla^\top z_{1,t}({\xi}_t) \\
                \vdots \\
                \nabla^\top z_{k,t}({\xi}_t)  \\
            \end{bmatrix}^\top
            \begin{bmatrix}
            \log \ell_{1,t} - \log \ell_{1,t+1} \\
            \vdots \\
            \log \ell_{k,t} - \log \ell_{k,t+1}.
            \end{bmatrix}.
            $$
        \ENDFOR
    \end{algorithmic}
    \label{alg:famo}
\end{algorithm*}

\subsection{Balanced Rate of Loss Improvement}
\label{sec::balanced-rate}
At time $t$, assume we perform the update $\x_{t+1} = \x_t - \alpha d_t$, we define the rate of improvement for task $i$ as
\begin{equation}
r_i(\alpha, d_t) = \frac{\ell_{i,t} - \ell_{i,t+1} }{\ell_{i,t}}.\footnote{To avoid division by zero, in practice we add a small constant (e.g., $1e-8$) to all losses. For the ease of notation (e.g., $\ell_i(\cdot) \leftarrow \ell_i(\cdot) + 1e-8$, we omit it throughout the paper.}
\end{equation}
\famo{} then seeks an update $d_t$ that results in the largest \emph{worst-case improvement rate} across all tasks ($\frac{1}{2}\norm{d_t}$ is subtracted to prevent an under-specified optimization problem where the objective can be infinitely large):
\begin{equation}
    \max_{d_t \in \RR^m} \min_{i \in [k]} \frac{1}{\alpha}r_i(\alpha, d_t) - \frac{1}{2}\norm{d_t}^2.
    \label{eq:original-primal}
\end{equation}
When the step size $\alpha$ is small, using Taylor approximation, the problem \eqref{eq:original-primal} 
can be approximated by  %transforms into
\begin{equation}
    \max_{d_t \in \RR^m} \min_{i \in [K]} \frac{ \nabla \ell_{i,t}^\top d_t}{\ell_{i,t}} - \frac{1}{2}\norm{d_t}^2 = \big(\nabla \log \ell_{i,t}\big)^\top d_t - \frac{1}{2}\norm{d_t}^2.
    \label{eq:primal}
\end{equation}
Instead of solving the primal problem in \eqref{eq:primal} where $d \in \RR^m$ ($m$ can be millions if $\x$ is the parameter of a neural network), we consider its dual problem:
\begin{pro}
    The dual objective of \eqref{eq:primal} is
\begin{equation}
    z_t^* \in \argmin_{z \in \S_k} \frac{1}{2}\norm{
     J_t z}^2,~~~\text{where}~~~J_t = \begin{bmatrix}
        \nabla \log \ell_{1,t}^\top \\
        \vdots \\
        \nabla \log \ell_{k,t}^\top
    \end{bmatrix},
    \label{eq:dual}
\end{equation}
where $z_t^* = [z_{t,i}^*]$ is the 
optimal combination weights of the gradients,  
and the optimal update direction is $d_t^* = J_t z_t^*$.
\end{pro}
%%%%%%%%%%%%%%%%%%%%%%%%%%%% PROOF %%%%%%%%%%%%%%%%%%%%%%%%%%%%%%%
\begin{proof}
\begin{align*}
&\max_{d \in \RR^m} \min_{i \in [k]}  \big(\nabla \log \ell_{i,t}\big)^\top d - \frac{1}{2}\norm{d}^2 \\
    = & \max_{d \in \RR^m}  \min_{{z} \in \S_k} \big(\sum_{i=1}^k z_i \nabla \log \ell_{i,t} \big)^\top d - \frac{1}{2}\norm{d}^2 ~~~~~~~ \\ 
    %\red{\textcolor{magenta}{(strong duality)}} \\
    = &   \min_{{z} \in \S_k} \max_{d \in \RR^m} \big(\sum_{i=1}^k z_i \nabla \log \ell_{i,t} \big)^\top d - \frac{1}{2}\norm{d}^2 \qquad\textcolor{magenta}{\text{(strong duality)}}
\end{align*}
Write $g(d, {z}) = \big(\sum_{i=1}^k z_i \nabla \log \ell_{i,t} \big)^\top d - \frac{1}{2}\norm{d}^2$, then by setting
$$
\pdv{g}{d} = {0} \quad\Longrightarrow\quad d^* = \sum_{i=1}^k z_i \nabla \log \ell_{i,t}.
$$
Plugging in $d^*$ back, we have
$$
\max_{d \in \RR^m} \min_{i \in [k]}  \big(\nabla \log \ell_{i,t}\big)^\top d - \frac{1}{2}\norm{d}^2 = \min_{z \in \S_k}\frac{1}{2}\norm{\sum_{i=1}^k z_i \nabla \log \ell_{i,t}}^2 = \min_{z \in \S_k} \frac{1}{2}\norm{J_t z}^2.
$$
At the optimum, we have $d_t^* = J_t z_t^*$.
\end{proof}
The dual problem in \eqref{eq:dual} can be viewed as optimizing the log objective of the multiple gradient descent algorithm (MGDA)~\citep{desideri2012multiple,sener2018multi}. Similar to MGDA, \eqref{eq:dual} only involves a decision variable of dimension $k \ll m$. Furthermore, if the optimal combination weights $z^*_t$ is an interior point of $\S_k$, 
then the improvement rates $r_i(\alpha, d_t^*)$ of the different tasks $i$ equal, as we show in the following result.
\begin{pro}
Assume $\{\ell_i\}_{i=1}^k$ are smooth and the optimal weights $z^*_t$ in \eqref{eq:dual} is an interior point of $\S_k$, then 
$$
\forall ~i \neq j \in [k],\qquad 
r_i^*(d_t^*) = r_j^*(d_t^*),
$$
where $r_i^*(d_t^*) = \lim_{\alpha \rightarrow 0} 
\frac{1}{\alpha} r_i(\alpha, d_t^*).$
\label{pro:primal}
\end{pro}
\begin{proof}

Consider the Lagrangian form of \eqref{eq:dual}
\begin{equation}
\mathcal{L}({z}, \lambda, {\mu}) = \frac{1}{2}\norm{\sum_{i=1}^k z_i \nabla \log \ell_{i,t}}^2 + \lambda \big(\sum_{i=1}^k z_i - 1\big) - \sum_{i=1}^k \mu_i z_i,~~\text{where}~\forall i, \mu_i \geq 0.
\end{equation}
When ${z}^*$ reaches the optimum, we have $\partial \mathcal{L}({z}, \lambda, {\mu}) / \partial {z} = {0}$, recall that $d_t^* = J_t z_t^*$, then 
\begin{align*}
 {J}_t^\top {J}_t {z}^* = -{\mu} - \lambda,~~~\text{where}~~~{J}_t = \begin{bmatrix}
        \nabla \log \ell_{1,t}^\top \\
        \vdots \\
        \nabla \log \ell_{k,t}^\top
    \end{bmatrix} \quad \Longrightarrow \quad {J}_t^\top d_t^* = -({\mu} + \lambda).
\end{align*}
When ${z}_t^*$ is an interior point of $\S_k$, we know that ${\mu} = {0}$. Hence ${J}_t^\top d_t^* = -\lambda$. This means,
$$
\forall i \neq j, \qquad \lim_{\alpha \rightarrow 0} 
\frac{1}{\alpha} r_i(\alpha, d_t^*) = \nabla \log \ell_{i,t}^\top d_t^* = \nabla \log \ell_{j,t}^\top d_t^* = \lim_{\alpha \rightarrow 0} \frac{1}{\alpha} r_j(\alpha, d_t^*).
$$
\end{proof}
\subsection{Fast Approximation by Amortizing over Time}
\label{sec:fast-approx}
Instead of fully solving \eqref{eq:dual} at each optimization step, \famo{} performs a single-step gradient descent on ${z}$, which amortizes the computation over the optimization trajectory:
\begin{equation}
    {z}_{t+1} = {z}_t - \alpha_z \tilde{\delta},~~~~\text{where}~~ \tilde{\delta} = \nabla_{{z}} \frac{1}{2}\norm{\sum_{i=1}^k z_{i,t} \nabla \log \ell_{i,t}}^2 = {J}_t^\top {J}_t {z}_t.
    \label{eq:local-conflict-update1}
\end{equation}
But then, note that
\begin{equation}
    \frac{1}{\alpha} \begin{bmatrix}
        \log \ell_{1,t} - \log \ell_{1,t+1} \\
        \vdots \\
        \log \ell_{k,t} - \log \ell_{k,t+1}
    \end{bmatrix}\approx {J}_t^\top d_t = {J}_t^\top {J}_t {z}_t,
   \label{eq:famo-update1-approx}
\end{equation}
so we can use the change in log losses to approximate the gradient.

In practice, to ensure that ${z}$ always stays in $\S_k$, we re-parameterize ${z}$ by ${\xi}$ and let ${z}_t = \textbf{Softmax}({\xi}_t)$, where ${\xi}_t \in \RR^K$ are the unconstrained softmax logits. Consequently, we have the following approximate update on ${\xi}$ from \eqref{eq:local-conflict-update1}:
\begin{equation}
    {\xi}_{t+1}  = {\xi}_t - \beta \delta,~~~~\text{where}~~ \delta = \begin{bmatrix}
                \nabla^\top z_{1,t}({\xi}) \\
                \vdots \\
                \nabla^\top z_{k,t}({\xi})  \\
    \end{bmatrix}^\top \begin{bmatrix}
        \log \ell_{1,t} - \log \ell_{1,t+1} \\
        \vdots \\
        \log \ell_{k,t} - \log \ell_{k,t+1}
    \end{bmatrix}.
    \label{eq:famo-1-update}
\end{equation}
\textbf{Remark:} While it is possible to perform gradient descent on ${z}$ for other gradient manipulation methods in principle, we will demonstrate in Appendix~\ref{sec::apx-fast-approx} that not all such updates can be easily approximated using the change in losses.

\subsection{Practical Implementation} 
To facilitate practical implementation, we present two modifications to the update in \eqref{eq:famo-1-update}.
\paragraph{Re-normalization} The suggested update above is a convex combination of the gradients of the log loss, e.g., 
$$d^* = \sum_{i=1}^k z_{i,t} \nabla \log \ell_{i,t} = \sum_{i=1}^k \big(\frac{z_{i,t}}{\ell_{i,t}}\big) \nabla \ell_{i,t}.$$
When $\ell_{i,t}$ is small, the multiplicative coefficient $\frac{z_{i,t}}{\ell_{i,t}}$ can be quite large and result in unstable optimization. Therefore, we propose to multiply $d^*$ by a constant $c_t$, such that $c_t d^*$ can be written as a convex combination of the task gradients just as in other gradient manipulation algorithms (see \eqref{eq:gradmanip} and we provide the corresponding definition of ${w}$ in the following):
\begin{equation}
c_t = \big(\sum_{i=1}^k \frac{z_{i,t}}{\ell_{i,t}}\big)^{-1}~~~~\text{and}~~~~d_t = c_t d^* = \sum_{i=1}^k w_i \nabla \ell_{i,t},~~~\text{where}~~w_i = c_t\frac{z_{i,t}}{\ell_{i,t}}.
\label{eq:famo-ct}
\end{equation}

\paragraph{Regularization} As we are amortizing the computation over time and the loss objective $\{\ell_i(\cdot)\}$s are changing dynamically, it makes sense to focus more on the recent updates of ${\xi}$~\cite{zhou2022convergence}. To this end, we put a decay term on ${w}$ such that the resulting ${\xi}_t$ is an exponential moving average of its gradient updates:
\begin{equation}
    {\xi}_{t+1} = {\xi}_t - \beta (\delta_t + \gamma {\xi}_t) = -\beta \big(\delta_t + (1-\beta \gamma) \delta_{t-1} + (1 - \beta \gamma)^2 \delta_{t-2} + \dots\big).
    \label{eq:famo-2-update}
\end{equation}

We provide the complete \famo{} algorithm in Algorithm~\ref{alg:famo} and its pseudocode in Appendix~\ref{sec::apx-pseudocode}.

\subsection{The Continuous Limit of \famo{}}
One way to characterize \famo{}'s behavior is to understand the stationary points of the continuous-time limit of \famo{} (i.e. when step sizes $(\alpha, \beta)$ shrink to zero). From Algorithm~\ref{alg:famo}, one can derive the following non-autonomous dynamical system (assuming $\{\ell_i\}$ are all smooth):
\begin{equation}
 \begin{bmatrix}
 ~\dot{\x}~ \\
 ~\dot{{\xi}}~
 \end{bmatrix} = -c_t \begin{bmatrix}
     {J}_t {z}_t\\
      {A}_t {J}_t^\top {J}_t {z}_t + \frac{\gamma}{c_t} {\xi}_t
    \end{bmatrix},~~\text{where}~{A}_t = \begin{bmatrix}
                \nabla^\top z_{1,t}({\xi}_t) \\
                \vdots \\
                \nabla^\top z_{k,t}({\xi}_t)  \\
    \end{bmatrix}.
    \label{eq:ode}
\end{equation}
\eqref{eq:ode} reaches its stationary points (or fixed points) when (note that $c_t > 0$)
\begin{equation}
    \begin{bmatrix}
 ~\dot{\x}~ \\
 ~\dot{{\xi}}~
 \end{bmatrix} = 0 ~~ \Longrightarrow ~~ {J}_t {z}_t = 0~~\text{and}~~
    {\xi}_t = 0 ~~ \Longrightarrow ~~ \sum_{i=1}^k \nabla \log \ell_{i,t} = 0.
\end{equation}
Therefore, the minimum points of $\sum_{i=1}^k \log \ell_i(\x)$ are all stationary points of \eqref{eq:ode}.

\section{Related Work}
\label{sec::related}
In this section, we summarize existing methods that tackle learning challenges in multitask learning (MTL). The general idea of most existing works is to encourage positive knowledge transfer by sharing parameters while decreasing any potential negative knowledge transfer (a.k.a, interference) during learning. There are three major ways of doing so: task grouping, designing network architectures specifically for MTL, and designing multitask optimization methods.

\paragraph{Task Grouping} Task grouping refers to grouping $K$ tasks into $N < K$ clusters and learning $N$ models for each cluster. The key is estimating the amount of positive knowledge transfer incurred by grouping certain tasks together and then identifying which tasks should be grouped~\citep{thrun1996discovering, zamir2018taskonomy, standley2020tasks, shen2021variational, fifty2021efficiently}.

\paragraph{Multitask Architecture} Novel neural architectures for MTL include \emph{hard-parameter-sharing} methods, which decompose a neural network into task-specific modules and a shared feature extractor using manually designed heuristics~\citep{kokkinos2017ubernet, long2017learning, bragman2019stochastic}, and \emph{soft-parameter-sharing} methods, which learn which parameters to share~\citep{misra2016cross, ruder2019latent, gao2020mtl, liu2019end}. Recent studies extend neural architecture search for MTL by learning where to branch a network to have task-specific modules~\cite{guo2020learning, bruggemann2020automated}.

\paragraph{Multitask Optimization} The most relevant approach to our method is MTL optimization via task balancing. These methods dynamically re-weight all task losses to mitigate the conflicting gradient issue~\cite{vandenhende2021multi, yu2020gradient}. The simplest form of gradient manipulation is to re-weight the task losses based on manually designed criteria~\cite{chen2018gradnorm, guo2018dynamic, kendall2018multi}, but these methods are often heuristic and lack theoretical support. Gradient manipulation methods~\cite{sener2018multi, yu2020gradient, liu2020towards, chen2020just, javaloy2021rotograd, liu2021conflict, navon2022multi,liu2022auto,zhugradient} propose to form a new update vector at each optimization by linearly combining task gradients. The local improvements across all tasks using the new update can often be explicitly analyzed, making these methods better understood in terms of convergence. However, it has been observed that gradient manipulation methods are often slow in practice, which may outweigh their performance benefits~\cite{kurin2022defense}. By contrast, \famo{} is designed to match the performance of these methods while remaining efficient in terms of memory and computation. Another recent work proposes to sample random task weights at each optimization step for MTL~\cite{lin2021closer}, which is also computationally efficient. However, we will demonstrate empirically that \famo{} performs better than this method.

\section{Empirical Results}
\label{sec::exp}
We conduct experiments to answer the following question:

% \noindent \qone{}: 
\emph{How does \famo{} perform in terms of space/time complexities and standard MTL metrics against prior MTL optimizers on standard benchmarks (e.g., supervised and reinforcement MTL problems)?}

% \noindent \qtwo{}: How does \famo{} behave with respect to the regularization coefficient $\gamma$?
% \qq{this questions feels a bit small and detailed to put here in the begining of the whole experiment section. how about the other hyperparameters other than $\gamma$? why is $\gamma$ so important?}

In the following, we first use a toy 2-task problem to demonstrate how \famo{} mitigates CG while being efficient. Then we show that \famo{} performs comparably or even better than state-of-the-art gradient manipulation methods on standard multitask supervised and reinforcement learning benchmarks. In addition, \famo{} requires significantly lower computation time when $K$ is large compared to other methods. Lastly, we conduct an ablation study on how robust \famo{} is to $\gamma$. Each subsection first details the experimental setup and then analyzes the results.

\subsection{A Toy 2-Task Example}
\label{sec::exp-toy}

\begin{figure}[h!]
    \centering
    \includegraphics[width=0.8\textwidth]{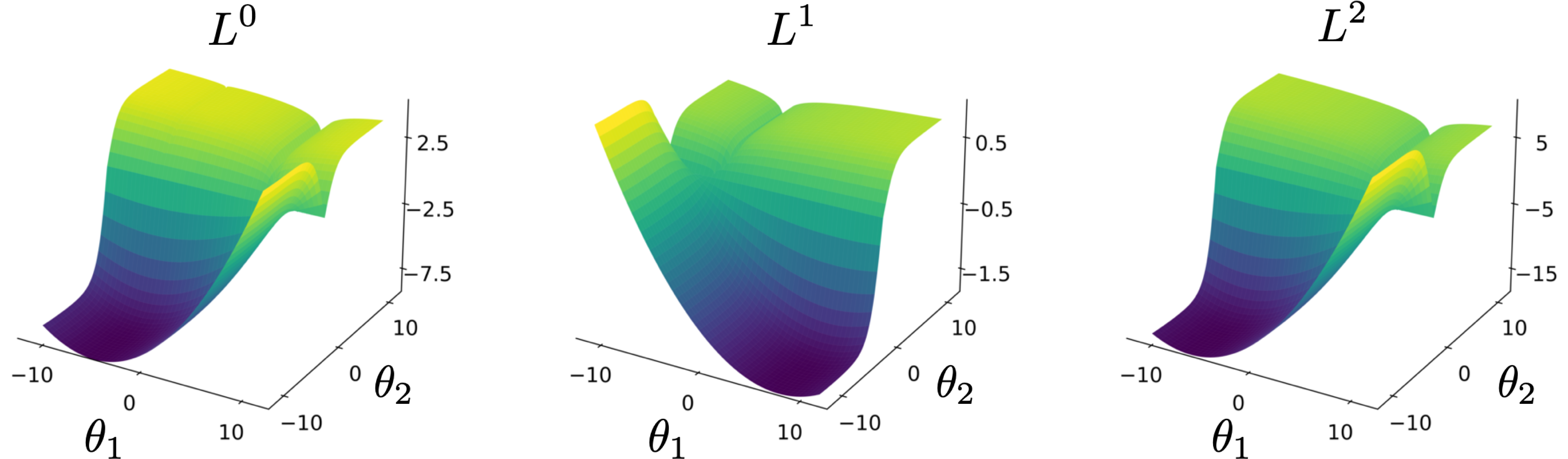}
    \caption{The average loss $L^0$ and the two task losses $L^1$ and $L^2$ for the toy example.}
    \label{fig:toy_plot}
\end{figure}
To better understand the optimization trajectory of \famo{}, we adopt the same 2D multitask optimization problem from \nashmtl{}~\citep{navon2022multi} to visualize how \famo{} balances different loss objectives. The model parameter $\theta=(\theta_1, \theta_2) \in \RR^2$. The two tasks' objectives and their surface plots are provided in Appendix~\ref{sec::apx-toy} and Figure~\ref{fig:toy_plot}. We compare \famo{} against \textsc{Adam}~\cite{kingma2014adam}, \mgda{}~\cite{sener2018multi}, \pcgrad{}~\cite{yu2020gradient}, \cagrad{}~\cite{liu2021conflict}, and \nashmtl{}~\cite{navon2022multi}.
We then pick 5 initial points $\theta_\text{init}\in\{(-8.5, 7.5), (-8.5, 5), (0,0), (9,9), (10, -8)\}$ and plot the corresponding optimization trajectories with different methods in Figure~\ref{fig:toy}. Note that the toy example is constructed such that naively applying \textsc{Adam} on the average loss can cause the failure of optimization for task 1.

\analysis{} From Figure~\ref{fig:toy}, we observe that \famo{}, like all other gradient manipulation methods, mitigates the CG and reaches the Pareto front for all five runs. In the meantime, \famo{} performs similarly to \nashmtl{} and achieves a balanced loss decrease even when the two task losses are improperly scaled. Finally, as shown in the top-right of the plot, \famo{} behaves similarly to \textsc{Adam} in terms of the training time, which is 25$\times$ faster than \nashmtl{}.

\subsection{MTL Performance}
\paragraph{Multitask Supervised Learning.} We consider four supervised benchmarks commonly used in prior MTL research~\cite{liu2021conflict,liu2019end,navon2022multi,pascal2021improved}: NYU-v2~\cite{SilbermanECCV12} (3 tasks), CityScapes~\cite{cordts2016cityscapes} (2 tasks), QM-9~\cite{blum} (11 tasks), and CelebA~\cite{liu2015faceattributes} (40 tasks). Specifically, NYU-v2 is an indoor scene dataset consisting of 1449 RGBD images and dense per-pixel labeling with 13 classes. The learning objectives include image segmentation, depth prediction, and surface normal prediction based on any scene image. CityScapes dataset is similar to NYU-v2 but contains 5000 street-view RGBD images with per-pixel annotations. QM-9 dataset is a widely used benchmark in graph neural network learning. It consists of $>$130K molecules represented as graphs annotated with node and edge features. We follow the same experimental setting used in \nashmtl{}~\cite{navon2022multi}, where the learning objective is to predict 11 properties of molecules. We use 110K molecules from the QM9 example in PyTorch Geometric~\cite{fey2019fast}, 10K molecules for validation, and the rest of 10K molecules for testing. The characteristic of this dataset is that the 11 properties are at different scales, posing a challenge for task balancing in MTL. Lastly, CelebA dataset contains 200K face images of 10K different celebrities, and each face image is provided with 40 facial binary attributes. Therefore, CelebA can be viewed as a 40-task MTL problem. Different from NYU-v2, CityScapes, and QM-9, the number of tasks ($K$) in CelebA is much larger, hence posing a challenge to learning efficiency.

We compare \famo{} against 11 MTL optimization methods and a single-task learning baseline: \textbf{(1)} Single task learning (\stl{}), training an independent model ($\theta$ for each task; \textbf{(2)} Linear scalarization (\ls{}) baseline that minimizes $L^0$; \textbf{(3)} Scale-invariant (\si{}) baseline that minimizes $\sum_k \log L^k(\theta)$, as SI is invariant to any scalar multiplication of task losses; \textbf{(4)} Dynamic Weight Average (\dwa{})~\cite{liu2019end}, a heuristic for adjusting task weights based on rates of loss changes; \textbf{(5)} Uncertainty Weighting (\uw{})~\cite{kendall2018multi} uses task uncertainty as a proxy to adjust task weights; \textbf{(6)} Random Loss Weighting (\rlw{})~\cite{lin2021closer} that samples task weighting whose log-probabilities follow the normal distribution; \textbf{(7)} \mgda{}~\cite{sener2018multi} that finds the equal descent direction for each task; \textbf{(8)} \pcgrad{}~\cite{yu2020gradient} proposes to project each task gradient to the normal plan of that of other tasks and combining them together in the end; \textbf{(9)} \cagrad{}~\cite{liu2021conflict} optimizes the average loss while explicitly controls the minimum decrease across tasks; \textbf{(10)} \imtlg{}~\cite{liu2020towards} finds the update direction with equal projections on task gradients; \textbf{(11)} \graddrop{}~\cite{chen2020just} that randomly dropout certain dimensions of the task gradients based on how much they conflict; \textbf{(12)} \nashmtl{}~\cite{navon2022multi} formulates MTL as a bargaining game and finds the solution to the game that benefits all tasks. For \famo{}, we choose the best hyperparameter $\gamma \in \{0.0001, 0.001, 0.01\}$ based on the validation loss. Specifically, we choose $\gamma$ equals $0.01$ for the CityScapes dataset and $0.001$ for the rest of the datasets. See Appendix~\ref{sec::apx-result-error} for results with error bars.

\begin{table*}[t!]
    \centering
    \resizebox{\textwidth}{!}{%
    \begin{tabular}{lrrrrrrrrrrr}
    \toprule
      &  \multicolumn{2}{c}{Segmentation} & \multicolumn{2}{c}{Depth} & \multicolumn{5}{c}{Surface Normal} & &\\
    \cmidrule(lr){2-3}\cmidrule(lr){4-5}\cmidrule(lr){6-10}
    \textbf{Method} &  \multirow{2}{*}{mIoU $\uparrow$} & \multirow{2}{*}{Pix Acc $\uparrow$} & \multirow{2}{*}{Abs Err $\downarrow$} & \multirow{2}{*}{Rel Err $\downarrow$} & \multicolumn{2}{c}{Angle Dist $\downarrow$} & \multicolumn{3}{c}{Within $t^\circ$ $\uparrow$}  & \mr{} $\downarrow$ &  \dm{} $\downarrow$ \\
    \cmidrule(lr){6-7}\cmidrule(lr){8-10}
    & & & & & Mean & Median & 11.25 & 22.5 & 30  &\\
    \midrule 
    \stl{}       & 38.30 & 63.76 & 0.6754 & 0.2780 & 25.01 & 19.21 & 30.14 & 57.20 & 69.15   &  &     \\
    \midrule
    \ls{}        & 39.29 & 65.33 & 0.5493 & 0.2263 & 28.15 & 23.96 & 22.09 & 47.50 & 61.08   & 8.89 & 5.59  \\
    \si{}        & 38.45 & 64.27 & 0.5354 & 0.2201 & 27.60 & 23.37 & 22.53 & 48.57 & 62.32   & 7.89 & 4.39  \\
    \rlw{}       & 37.17 & 63.77 & 0.5759 & 0.2410 & 28.27 & 24.18 & 22.26 & 47.05 & 60.62   &11.22 & 7.78  \\
    \dwa{}       & 39.11 & 65.31 & 0.5510 & 0.2285 & 27.61 & 23.18 & 24.17 & 50.18 & 62.39   & 7.67 & 3.57  \\
    \uw{}        & 36.87 & 63.17 & 0.5446 & 0.2260 & 27.04 & 22.61 & 23.54 & 49.05 & 63.65   & 7.44 & 4.05  \\
    \mgda{}      & 30.47 & 59.90 & 0.6070 & 0.2555 & \best{24.88} & \best{19.45} & 29.18 & \best{56.88} & \best{69.36}   & 6.00 & 1.38  \\
    \pcgrad{}    & 38.06 & 64.64 & 0.5550 & 0.2325 & 27.41 & 22.80 & 23.86 & 49.83 & 63.14   & 8.00 & 3.97  \\
    \graddrop{}  & 39.39 & 65.12 & 0.5455 & 0.2279 & 27.48 & 22.96 & 23.38 & 49.44 & 62.87   & 7.00 & 3.58  \\
    \cagrad{}    & 39.79 & 65.49 & 0.5486 & 0.2250 & 26.31 & 21.58 & 25.61 & 52.36 & 65.58   & 4.56 & 0.20  \\
    \imtlg{}     & 39.35 & 65.60 & 0.5426 & 0.2256 & 26.02 & 21.19 & 26.20 & 53.13 & 66.24   & 3.78 & -0.76  \\
    \nashmtl{}   & \best{40.13} & \best{65.93} & \best{0.5261} & \best{0.2171} & 25.26 & 20.08 & 28.40 & 55.47 & 68.15   & \best{2.11} & -4.04  \\
    \midrule
    \famo{}      & 38.88 & 64.90 & 0.5474 & 0.2194 & 25.06 & 19.57 & \best{29.21} & 56.61 & 68.98   & 3.44 & \best{-4.10} \\
    \bottomrule  
    \end{tabular}
    }
    \vspace{2pt}
    \caption{Results on NYU-v2 dataset (3 tasks). Each experiment is repeated over 3 random seeds and the mean is reported. The best average result is marked in bold. \mr{} and \dm{} are the main metrics for MTL performance.}
    \label{tab:nyu-v2}
    \vspace{-10pt}
\end{table*}

\begin{table*}[t!]
    \centering
    \resizebox{\textwidth}{!}{%
    \begin{tabular}{lrrrrrrrrrrrrr}
    \toprule
    \multirow{2}{*}{\textbf{Method}} & $\mu$ & $\alpha$ & $\epsilon_\text{HOMO}$ & $\epsilon_\text{LUMO}$ & $\langle R^2\rangle$ & ZPVE & $U_0$ & $U$ & $H$ & $G$ & $c_v$ &\multirow{2}{*}{\mr{} $\downarrow$} & \multirow{2}{*}{ \dm{} $\downarrow$}\\
      \cmidrule(lr){2-12}
      & \multicolumn{11}{c}{MAE $\downarrow$} & & \\
    \midrule 
        \stl{}      & 0.07 & 0.18 &  60.6 &  53.9 & 0.50  &  4.53  &  58.8  &  64.2 &  63.8 &  66.2 & 0.07 &     &       \\
        \midrule
        \ls{}       & 0.11 & 0.33 &  \best{73.6} &  89.7 & 5.20  & 14.06  & 143.4  & 144.2 & 144.6 & 140.3 & 0.13 & 6.45 & 177.6 \\
        \si{}       & 0.31 & 0.35 & 149.8 & 135.7 & \best{1.00}  &  \best{4.51}  &  \best{55.3}  &  \best{55.8} &  \best{55.8} &  \best{55.3} & 0.11 & 3.55 &  77.8 \\
        \rlw{}      & 0.11 & 0.34 &  76.9 &  92.8 & 5.87  & 15.47  & 156.3  & 157.1 & 157.6 & 153.0 & 0.14 & 8.00 & 203.8 \\
        \dwa{}      & 0.11 & 0.33 &  74.1 &  90.6 & 5.09  & 13.99  & 142.3  & 143.0 & 143.4 & 139.3 & 0.13 & 6.27 & 175.3 \\
        \uw{}       & 0.39 & 0.43 & 166.2 & 155.8 & 1.07  &  4.99  &  66.4  &  66.8 &  66.8 &  66.2 & 0.12 & 4.91 & 108.0 \\
        \mgda{}     & 0.22 & 0.37 & 126.8 & 104.6 & 3.23  &  5.69  &  88.4  &  89.4 &  89.3 &  88.0 & 0.12 & 5.91 & 120.5 \\
        \pcgrad{}   & 0.11 & 0.29 &  75.9 &  88.3 & 3.94  &  9.15  & 116.4  & 116.8 & 117.2 & 114.5 & 0.11 & 4.73 & 125.7 \\
        \cagrad{}   & 0.12 & 0.32 &  83.5 &  94.8 & 3.22  &  6.93  & 114.0  & 114.3 & 114.5 & 112.3 & 0.12 & 5.45 & 112.8 \\
        \imtlg{}    & 0.14 & 0.29 &  98.3 &  93.9 & 1.75  &  5.70  & 101.4  & 102.4 & 102.0 & 100.1 & 0.10 & 4.36 &  77.2 \\
        \nashmtl{}  & \best{0.10} & \best{0.25} &  82.9 &  \best{81.9} & 2.43  &  5.38  &  74.5  &  75.0 &  75.1 &  74.2 & \best{0.09} & 2.09 &  62.0 \\
    \midrule 
    \famo{}     & 0.15 & 0.30 & 94.0 & 95.2 &  1.63 & 4.95 & 70.82 & 71.2 & 71.2 & 70.3 & 0.10 & 3.27 & \best{58.5} \\
    \bottomrule 
    \end{tabular}
    }
    \vspace{2pt}
    \caption{Results on QM-9 dataset (11 tasks). Each experiment is repeated over 3 random seeds and the mean is reported. The best average result is marked in bold. \mr{} and \dm{} are the main metrics for MTL performance.}
    \label{tab:qm9}
    \vspace{-10pt}
\end{table*}
\begin{table*}[t!]
    \centering
    \resizebox{\textwidth}{!}{%
    \begin{tabular}{lrrrrrrrr}
    \toprule
   \multirow{4}{*}{\textbf{Method}} & \multicolumn{6}{c}{\textbf{CityScapes}} & \multicolumn{2}{c}{\textbf{CelebA}} \\
      \cmidrule(lr){2-7}\cmidrule(lr){8-9}
      &  \multicolumn{2}{c}{Segmentation} & \multicolumn{2}{c}{Depth} & \multirow{2}{*}{\mr{} $\downarrow$} & \multirow{2}{*}{ \dm{} $\downarrow$}  & \multirow{2}{*}{\mr{} $\downarrow$} & \multirow{2}{*}{ \dm{} $\downarrow$}\\
    \cmidrule(lr){2-3}\cmidrule(lr){4-5}
     &  mIoU $\uparrow$ & Pix Acc $\uparrow$ & Abs Err $\downarrow$ & Rel Err $\downarrow$ &  &  & &\\
    \midrule 
    \stl{}      & 74.01 & 93.16 & 0.0125 & 27.77 & & \\
    \midrule
    \ls{}       & 70.95 & 91.73 & 0.0161 & 33.83 & 6.50 & 14.11 & 4.15 & 6.28 \\
    \si{}       & 70.95 & 91.73 & 0.0161 & 33.83 & 9.25 & 14.11 & 7.20 & 7.83 \\
    \rlw{}      & 74.57 & 93.41 & 0.0158 & 47.79 & 9.25 & 24.38 & 1.46 & 5.22 \\
    \dwa{}      & 75.24 & 93.52 & 0.0160 & 44.37 & 6.50 & 21.45 & 3.20 & 6.95 \\
    \uw{}       & 72.02 & 92.85 & 0.0140 & \best{30.13} & 6.00 & \best{5.89} & 3.23 & 5.78\\
    \mgda{}     & 68.84 & 91.54 & 0.0309 & 33.50 & 9.75 & 44.14 & 14.85 & 10.93 \\
    \pcgrad{}   & 75.13 & 93.48 & 0.0154 & 42.07 & 6.75 & 18.29 & 3.17 & 6.65 \\
    \graddrop{} & 75.27 & 93.53 & 0.0157 & 47.54 & 6.00 & 23.73 & 3.29 & 7.80 \\
    \cagrad{}   & 75.16 & 93.48 & 0.0141 & 37.60 & 5.75 & 11.64 & 2.48 & 6.20 \\
    \imtlg{}    & 75.33 & 93.49 & 0.0135 & 38.41 & 4.00 & 11.10 & \best{0.84} & \best{4.67} \\
    \nashmtl{}  & \best{75.41} & \best{93.66} & \best{0.0129} & 35.02 & \best{2.00} & 6.82 & 2.84 & 4.97 \\
    \midrule 
    \famo{}     & 74.54 & 93.29 & 0.0145 & 32.59 & 6.25  & 8.13 & 1.21 & 4.72 \\
    \bottomrule 
    \end{tabular}
    }
    \vspace{2pt}
    \caption{Results on CityScapes (2 tasks) and CelebA (40 tasks) datasets. Each experiment is repeated over 3 random seeds and the mean is reported. The best average result is marked in bold. \mr{} and \dm{} are the main metrics for MTL performance.}
    \label{tab:cityscapes-and-celeba}
    \vspace{-20pt}
\end{table*}

\evaluations{} We consider two metrics~\cite{navon2022multi} for MTL:
\textbf{1)} \dm{}, the average per-task performance drop of a method $m$ relative to the \stl{} baseline denoted as $b$:
    $
        \bm{\Delta}\bm{m}\% = \frac{1}{K}\sum_{k=1}^K (-1)^{\delta_k} (M_{m,k} - M_{b,k}) / M_{b,k} \times 100,
    $
    where $M_{b,k}$ and $M_{m,k}$ are the STL and $m$'s value for metric $M_k$. $\delta_k = 1$ (or $0$) if the $M_k$ is higher (or lower) the better. 
\textbf{2) Mean Rank (MR)}: the average rank of each method across tasks. For instance, if a method ranks first for every task, \textbf{MR} will be 1.
%\end{itemize}

\analysis{} Results on the four benchmark datasets are provided in Table~\ref{tab:nyu-v2},~\ref{tab:qm9} and~\ref{tab:cityscapes-and-celeba}. We observe that \famo{} performs consistently well across different supervised learning MTL benchmarks compared to other gradient manipulation methods. In particular, it achieves state-of-the-art results in terms of \dm{} on the NYU-v2 and QM-9 datasets.

\paragraph{Multitask Reinforcement Learning.} We further apply \famo{} to multitask reinforcement learning (MTRL) problems as MTRL often suffers more from conflicting gradients due to the stochastic nature of reinforcement learning~\cite{yu2020gradient}. Following \cagrad{}~\cite{liu2021conflict}, we apply \famo{} on the MetaWorld~\cite{yu2020meta} MT10 benchmark, which consists of 10 robot manipulation tasks with different reward functions. Following~\cite{sodhani2021multi}, we use Soft Actor-Critic (SAC)~\cite{haarnoja2018soft} as the underlying RL algorithm, and compare against baseline methods including \ls{} (SAC with a shared model)~\cite{yu2020meta}, Soft Modularization~\cite{yang2020multi} (an MTL network that routes different modules in a shared model to
form different policies), \pcgrad{}~\cite{yu2020gradient}, \cagrad{} and \nashmtl{}~\cite{navon2022multi}. The experimental setting and hyperparameters all match exactly with those in \cagrad{}. For \nashmtl{}, we report the results of applying the \nashmtl{} update once per $\{1, 50, 100\}$ iterations.\footnote{We could not reproduce the MTRL results of \nashmtl{} exactly, so we report both the results from the original paper and our reproduced results.} The results for all methods are provided in Table~\ref{tab:metaworld}.
% \begin{table}[h!]
%     \centering
%     \resizebox{0.5\textwidth}{!}{%
%     \begin{tabular}{lc}
%     \toprule
%     \multirow{2}{*}{\textbf{Method}} & Success \\
%     & (mean $\pm$ stderr)  \\
%     \midrule
%     \ls{} (lower bound)   & 0.49 \fs{0.07} \\
%     \stl{} (proxy for upper bound)              & 0.90 \fs{0.03} \\
%     \midrule
%     \pcgrad{}~\citep{yu2020gradient}                  & 0.72 \fs{0.02} \\
%     \textsc{Soft Modularization}~\citep{yang2020multi}& 0.73 \fs{0.04} \\
%     \cagrad{}                                         & 0.83 \fs{0.05} \\
%     \nashmtl{}~\cite{navon2022multi} (every 1)        & \best{0.91} \fs{0.03} \\
%     \nashmtl{}~\cite{navon2022multi} (every 50)       & 0.85 \fs{0.02} \\
%     \nashmtl{}~\cite{navon2022multi} (every 100)      & 0.87 \fs{0.03} \\
%     \nashmtl{} (ours) (every 1)                       & 0.80 \fs{0.13} \\
%     \nashmtl{} (ours) (every 50)                      & 0.76 \fs{0.10} \\
%     \nashmtl{} (ours) (every 100)                     & 0.80 \fs{0.12} \\
%     \uw{}~\cite{kendall2018multi}                     & 0.77 \fs{0.05} \\
%     \midrule
%     \famo{} (ours)                                    & 0.83 \fs{0.05} \\
%     \bottomrule
%     \end{tabular}
%     }
%     \vspace{10pt}
%     \caption{Multi-task reinforcement learning results on the Metaworld benchmarks. Results are average over 10 independent runs and the best result is marked in bold.}
%     \label{tab:metaworld}
% \end{table}

\begin{minipage}{\textwidth}
\begin{minipage}[b]{0.49\textwidth}
\centering
\vspace{10pt}
\includegraphics[width=\textwidth]{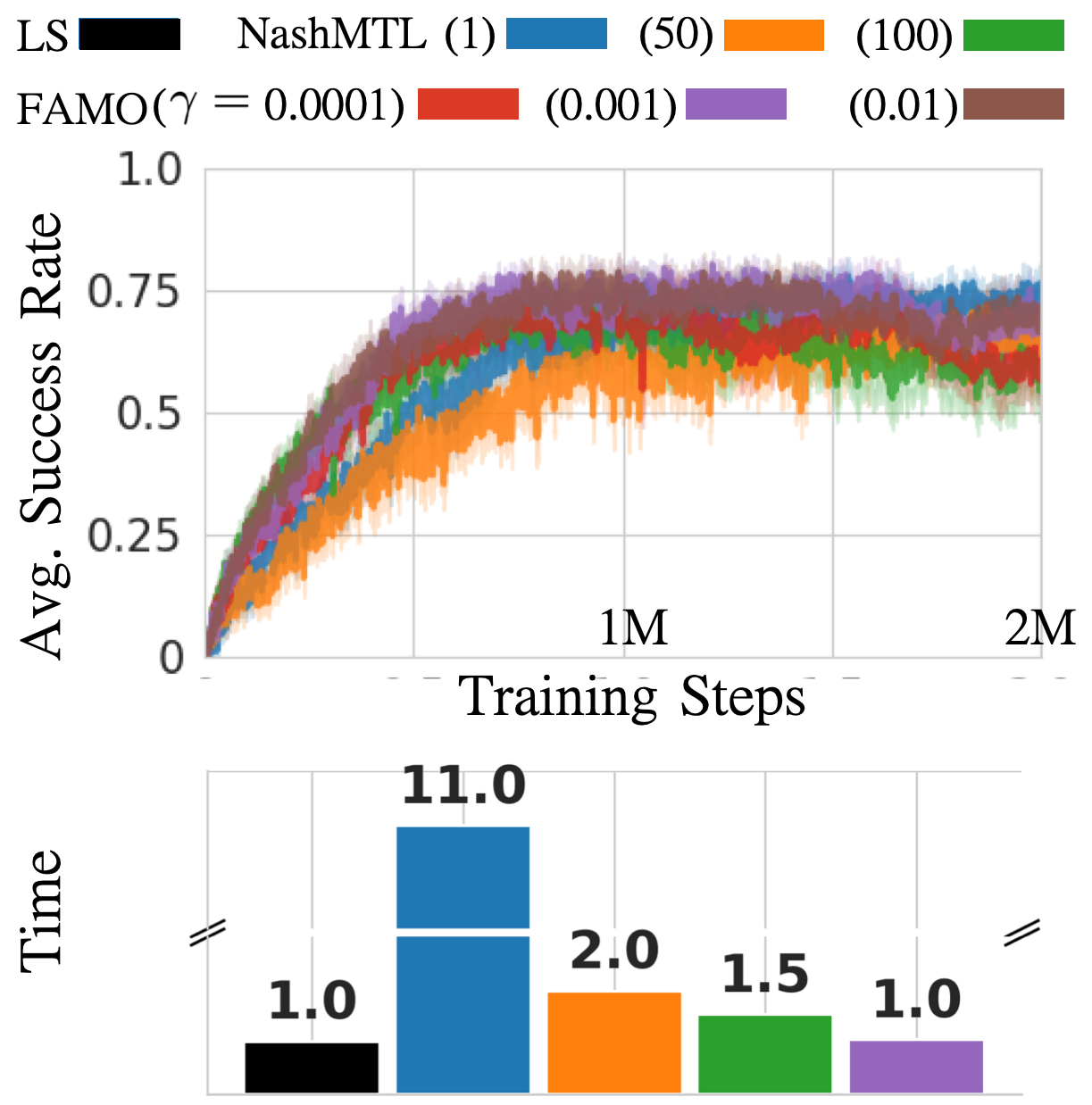}
\captionof{figure}{Training Success Rate and Time.}
\end{minipage}
\hfill
\begin{minipage}[b]{0.49\textwidth}
\centering
    \resizebox{\textwidth}{!}{%
    \begin{tabular}{lc}
    \toprule
    \multirow{2}{*}{\textbf{Method}} & Success $\uparrow$ \\
    & (mean $\pm$ stderr)  \\
    \midrule
    \ls{} (lower bound)   & 0.49 \fs{0.07} \\
    \stl{} (proxy for upper bound)              & 0.90 \fs{0.03} \\
    \midrule
    \pcgrad{}~\citep{yu2020gradient}                  & 0.72 \fs{0.02} \\
    \textsc{Soft Modularization}~\citep{yang2020multi}& 0.73 \fs{0.04} \\
    \cagrad{}                                         & 0.83 \fs{0.05} \\
    \nashmtl{}~\cite{navon2022multi} (every 1)        & \best{0.91} \fs{0.03} \\
    \nashmtl{}~\cite{navon2022multi} (every 50)       & 0.85 \fs{0.02} \\
    \nashmtl{}~\cite{navon2022multi} (every 100)      & 0.87 \fs{0.03} \\
    \midrule
    \nashmtl{} (ours) (every 1)                       & 0.80 \fs{0.13} \\
    \nashmtl{} (ours) (every 50)                      & 0.76 \fs{0.10} \\
    \nashmtl{} (ours) (every 100)                     & 0.80 \fs{0.12} \\
    \uw{}~\cite{kendall2018multi}                     & 0.77 \fs{0.05} \\
    \midrule
    \famo{} (ours)                                    & 0.83 \fs{0.05} \\
    \bottomrule
    \end{tabular}
    }
    \vspace{10pt}
    \label{tab:metaworld}
  \captionof{table}{MTRL results (averaged over 10 runs) on the Metaworld-10 benchmark.}
  \vspace{-5pt}
\end{minipage}
\vspace{10pt}
\end{minipage}
  
\analysis{} From Table~\ref{tab:metaworld}, we observe that \famo{} performs comparably to \cagrad{} and outperforms \pcgrad{} and the average gradient descent baselines by a large margin. \famo{} also outperforms \nashmtl{} based on our implementation. Moreover, \famo{} is significantly faster than \nashmtl{}, even when it is applied once every 100 steps.

\subsection{MTL Efficiency (Training Time Comparison)}
\begin{figure}[ht!]
    \centering
    \includegraphics[width=\textwidth]{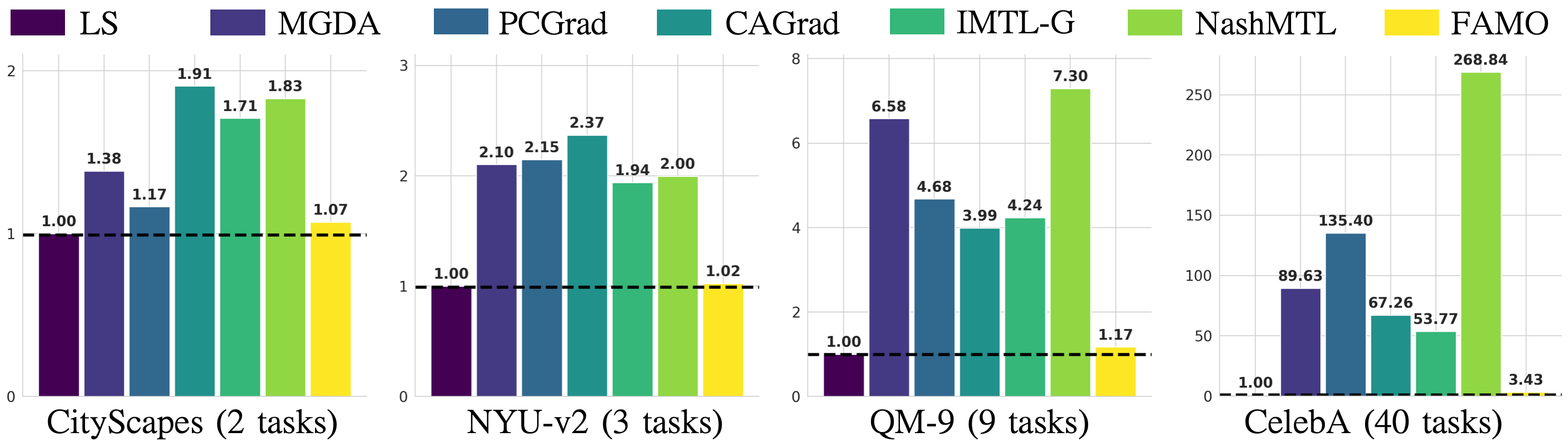}
    \caption{Average training time per epoch for different MTL optimization methods. We report the relative training time of a method to that of the linear scalarization (LS) method (which uses the average gradient).}
    \label{fig:time}
    \vspace{-10pt}
\end{figure}

Figure~\ref{fig:time} provides the \famo{}'s average training time per epoch against that of the baseline methods.

\analysis{} From the figure, we observe that \famo{} introduces negligible overhead across all benchmark datasets compared to the \ls{} method, which is, in theory, the lower bound for computation time. In contrast, methods like \nashmtl{} have much longer training time compared to \famo{}. More importantly, the computation cost of these methods scales with the number of tasks. In addition, note that these methods also take at least $\mathcal{O}(K)$ space to store the task gradients, which is implausible for large models in the many-task setting (i.e., when $m = |\theta|$ and $K$ are large).

\subsection{Ablation on $\gamma$}
\begin{figure}[ht!]
    \centering
    \includegraphics[width=\textwidth]{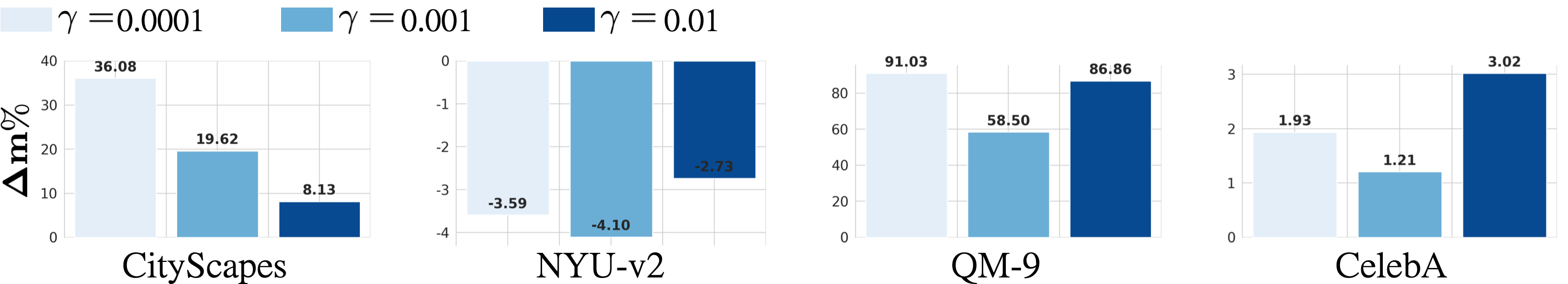}
    \caption{Ablation over $\gamma$: we plot the performance of \famo{} (in terms of \dm{} using different values of $\gamma$ from $\{0.0001,0.001, 0.01\}$ on the four supervised MTL benchmarks.}
    \label{fig:gamma}
        \vspace{-10pt}
\end{figure}

In this section, we provide the ablation study on the regularization coefficient $\gamma$ in Figure~\ref{fig:gamma}.

\analysis{} From Figure~\ref{fig:gamma}, we can observe that choosing the right regularization coefficient can be crucial. But except for CityScapes, \famo{} performs reasonably well using all different $\gamma$s. The problem with CityScapes is that one of the task losses is close to 0 at the very beginning, hence small changes in task weighting can result in very different loss improvement. Therefore we conjecture that using a larger $\gamma$, in this case, can help stabilize MTL.
\section{Conclusion and Limitations}
\label{sec::conclusion}
In this work, we introduce \famo{}, a fast optimization method for multitask learning (MTL) that mitigates the conflicting gradients using $\mathcal{O}(1)$ space and time. As multitasking large models gain more attention, we believe designing efficient but effective optimizers like \famo{} for MTL is crucial. \famo{} balances task losses by ensuring each task's loss decreases approximately at an equal rate. Empirically, we observe that \famo{} can achieve competitive performance against the state-of-the-art MTL gradient manipulation methods. One limitation of \famo{} is its dependency on the regularization parameter $\gamma$, which is introduced due to the stochastic update of the task weighting logits $\bm{w}$. Future work can investigate a more principled way of determining $\gamma$.

\clearpage

\bibliography{neurips_2023.bib}

\begin{thebibliography}{10}

\bibitem{blum}
L.~C. Blum and J.-L. Reymond.
\newblock 970 million druglike small molecules for virtual screening in the
  chemical universe database {GDB-13}.
\newblock {\em J. Am. Chem. Soc.}, 131:8732, 2009.

\bibitem{bragman2019stochastic}
Felix~JS Bragman, Ryutaro Tanno, Sebastien Ourselin, Daniel~C Alexander, and
  Jorge Cardoso.
\newblock Stochastic filter groups for multi-task cnns: Learning specialist and
  generalist convolution kernels.
\newblock In {\em Proceedings of the IEEE/CVF International Conference on
  Computer Vision}, pages 1385--1394, 2019.

\bibitem{bruggemann2020automated}
David Bruggemann, Menelaos Kanakis, Stamatios Georgoulis, and Luc Van~Gool.
\newblock Automated search for resource-efficient branched multi-task networks.
\newblock {\em arXiv preprint arXiv:2008.10292}, 2020.

\bibitem{bubeck2023sparks}
S{\'e}bastien Bubeck, Varun Chandrasekaran, Ronen Eldan, Johannes Gehrke, Eric
  Horvitz, Ece Kamar, Peter Lee, Yin~Tat Lee, Yuanzhi Li, Scott Lundberg,
  et~al.
\newblock Sparks of artificial general intelligence: Early experiments with
  gpt-4.
\newblock {\em arXiv preprint arXiv:2303.12712}, 2023.

\bibitem{caruana1997multitask}
Rich Caruana.
\newblock Multitask learning.
\newblock {\em Machine learning}, 28(1):41--75, 1997.

\bibitem{chen2018gradnorm}
Zhao Chen, Vijay Badrinarayanan, Chen-Yu Lee, and Andrew Rabinovich.
\newblock Gradnorm: Gradient normalization for adaptive loss balancing in deep
  multitask networks.
\newblock In {\em International Conference on Machine Learning}, pages
  794--803. PMLR, 2018.

\bibitem{chen2020just}
Zhao Chen, Jiquan Ngiam, Yanping Huang, Thang Luong, Henrik Kretzschmar, Yuning
  Chai, and Dragomir Anguelov.
\newblock Just pick a sign: Optimizing deep multitask models with gradient sign
  dropout.
\newblock {\em arXiv preprint arXiv:2010.06808}, 2020.

\bibitem{cordts2016cityscapes}
Marius Cordts, Mohamed Omran, Sebastian Ramos, Timo Rehfeld, Markus Enzweiler,
  Rodrigo Benenson, Uwe Franke, Stefan Roth, and Bernt Schiele.
\newblock The cityscapes dataset for semantic urban scene understanding.
\newblock In {\em Proceedings of the IEEE conference on computer vision and
  pattern recognition}, pages 3213--3223, 2016.

\bibitem{desideri2012multiple}
Jean-Antoine D{\'e}sid{\'e}ri.
\newblock Multiple-gradient descent algorithm (mgda) for multiobjective
  optimization.
\newblock {\em Comptes Rendus Mathematique}, 350(5-6):313--318, 2012.

\bibitem{fey2019fast}
Matthias Fey and Jan~Eric Lenssen.
\newblock Fast graph representation learning with pytorch geometric.
\newblock {\em arXiv preprint arXiv:1903.02428}, 2019.

\bibitem{fifty2021efficiently}
Chris Fifty, Ehsan Amid, Zhe Zhao, Tianhe Yu, Rohan Anil, and Chelsea Finn.
\newblock Efficiently identifying task groupings for multi-task learning.
\newblock {\em Advances in Neural Information Processing Systems},
  34:27503--27516, 2021.

\bibitem{gao2020mtl}
Yuan Gao, Haoping Bai, Zequn Jie, Jiayi Ma, Kui Jia, and Wei Liu.
\newblock Mtl-nas: Task-agnostic neural architecture search towards
  general-purpose multi-task learning.
\newblock In {\em Proceedings of the IEEE/CVF Conference on computer vision and
  pattern recognition}, pages 11543--11552, 2020.

\bibitem{guo2018dynamic}
Michelle Guo, Albert Haque, De-An Huang, Serena Yeung, and Li~Fei-Fei.
\newblock Dynamic task prioritization for multitask learning.
\newblock In {\em Proceedings of the European conference on computer vision
  (ECCV)}, pages 270--287, 2018.

\bibitem{guo2020learning}
Pengsheng Guo, Chen-Yu Lee, and Daniel Ulbricht.
\newblock Learning to branch for multi-task learning.
\newblock In {\em International Conference on Machine Learning}, pages
  3854--3863. PMLR, 2020.

\bibitem{haarnoja2018soft}
Tuomas Haarnoja, Aurick Zhou, Pieter Abbeel, and Sergey Levine.
\newblock Soft actor-critic: Off-policy maximum entropy deep reinforcement
  learning with a stochastic actor.
\newblock In {\em International Conference on Machine Learning}, pages
  1861--1870. PMLR, 2018.

\bibitem{javaloy2021rotograd}
Adri{\'a}n Javaloy and Isabel Valera.
\newblock Rotograd: Dynamic gradient homogenization for multi-task learning.
\newblock {\em arXiv preprint arXiv:2103.02631}, 2021.

\bibitem{katrutsa2020follow}
Alexandr Katrutsa, Daniil Merkulov, Nurislam Tursynbek, and Ivan Oseledets.
\newblock Follow the bisector: a simple method for multi-objective
  optimization.
\newblock {\em arXiv preprint arXiv:2007.06937}, 2020.

\bibitem{kendall2018multi}
Alex Kendall, Yarin Gal, and Roberto Cipolla.
\newblock Multi-task learning using uncertainty to weigh losses for scene
  geometry and semantics.
\newblock In {\em Proceedings of the IEEE conference on computer vision and
  pattern recognition}, pages 7482--7491, 2018.

\bibitem{kingma2014adam}
Diederik~P Kingma and Jimmy Ba.
\newblock Adam: A method for stochastic optimization.
\newblock {\em arXiv preprint arXiv:1412.6980}, 2014.

\bibitem{kirillov2023segment}
Alexander Kirillov, Eric Mintun, Nikhila Ravi, Hanzi Mao, Chloe Rolland, Laura
  Gustafson, Tete Xiao, Spencer Whitehead, Alexander~C Berg, Wan-Yen Lo, et~al.
\newblock Segment anything.
\newblock {\em arXiv preprint arXiv:2304.02643}, 2023.

\bibitem{kokkinos2017ubernet}
Iasonas Kokkinos.
\newblock Ubernet: Training a universal convolutional neural network for low-,
  mid-, and high-level vision using diverse datasets and limited memory.
\newblock In {\em Proceedings of the IEEE conference on computer vision and
  pattern recognition}, pages 6129--6138, 2017.

\bibitem{kurin2022defense}
Vitaly Kurin, Alessandro De~Palma, Ilya Kostrikov, Shimon Whiteson, and Pawan~K
  Mudigonda.
\newblock In defense of the unitary scalarization for deep multi-task learning.
\newblock {\em Advances in Neural Information Processing Systems},
  35:12169--12183, 2022.

\bibitem{lin2021closer}
Baijiong Lin, Feiyang Ye, and Yu~Zhang.
\newblock A closer look at loss weighting in multi-task learning.
\newblock {\em arXiv preprint arXiv:2111.10603}, 2021.

\bibitem{liu2021conflict}
Bo~Liu, Xingchao Liu, Xiaojie Jin, Peter Stone, and Qiang Liu.
\newblock Conflict-averse gradient descent for multi-task learning.
\newblock {\em Advances in Neural Information Processing Systems},
  34:18878--18890, 2021.

\bibitem{liu2020towards}
Liyang Liu, Yi~Li, Zhanghui Kuang, Jing-Hao Xue, Yimin Chen, Wenming Yang,
  Qingmin Liao, and Wayne Zhang.
\newblock Towards impartial multi-task learning.
\newblock In {\em International Conference on Learning Representations}, 2020.

\bibitem{liu2022auto}
Shikun Liu, Stephen James, Andrew~J Davison, and Edward Johns.
\newblock Auto-lambda: Disentangling dynamic task relationships.
\newblock {\em arXiv preprint arXiv:2202.03091}, 2022.

\bibitem{liu2019end}
Shikun Liu, Edward Johns, and Andrew~J Davison.
\newblock End-to-end multi-task learning with attention.
\newblock In {\em Proceedings of the IEEE/CVF Conference on Computer Vision and
  Pattern Recognition}, pages 1871--1880, 2019.

\bibitem{liu2015faceattributes}
Ziwei Liu, Ping Luo, Xiaogang Wang, and Xiaoou Tang.
\newblock Deep learning face attributes in the wild.
\newblock In {\em Proceedings of International Conference on Computer Vision
  (ICCV)}, December 2015.

\bibitem{long2017learning}
Mingsheng Long, Zhangjie Cao, Jianmin Wang, and Philip~S Yu.
\newblock Learning multiple tasks with multilinear relationship networks.
\newblock {\em Advances in neural information processing systems}, 30, 2017.

\bibitem{misra2016cross}
Ishan Misra, Abhinav Shrivastava, Abhinav Gupta, and Martial Hebert.
\newblock Cross-stitch networks for multi-task learning.
\newblock In {\em Proceedings of the IEEE conference on computer vision and
  pattern recognition}, pages 3994--4003, 2016.

\bibitem{SilbermanECCV12}
Pushmeet~Kohli Nathan~Silberman, Derek~Hoiem and Rob Fergus.
\newblock Indoor segmentation and support inference from rgbd images.
\newblock In {\em ECCV}, 2012.

\bibitem{navon2022multi}
Aviv Navon, Aviv Shamsian, Idan Achituve, Haggai Maron, Kenji Kawaguchi, Gal
  Chechik, and Ethan Fetaya.
\newblock Multi-task learning as a bargaining game.
\newblock {\em arXiv preprint arXiv:2202.01017}, 2022.

\bibitem{pascal2021improved}
Lucas Pascal, Pietro Michiardi, Xavier Bost, Benoit Huet, and Maria~A Zuluaga.
\newblock Improved optimization strategies for deep multi-task networks.
\newblock {\em arXiv preprint arXiv:2109.11678}, 2021.

\bibitem{ruder2019latent}
Sebastian Ruder, Joachim Bingel, Isabelle Augenstein, and Anders S{\o}gaard.
\newblock Latent multi-task architecture learning.
\newblock In {\em Proceedings of the AAAI Conference on Artificial
  Intelligence}, volume~33, pages 4822--4829, 2019.

\bibitem{sener2018multi}
Ozan Sener and Vladlen Koltun.
\newblock Multi-task learning as multi-objective optimization.
\newblock {\em arXiv preprint arXiv:1810.04650}, 2018.

\bibitem{shen2021variational}
Jiayi Shen, Xiantong Zhen, Marcel Worring, and Ling Shao.
\newblock Variational multi-task learning with gumbel-softmax priors.
\newblock {\em Advances in Neural Information Processing Systems},
  34:21031--21042, 2021.

\bibitem{sodhani2021multi}
Shagun Sodhani, Amy Zhang, and Joelle Pineau.
\newblock Multi-task reinforcement learning with context-based representations.
\newblock {\em arXiv preprint arXiv:2102.06177}, 2021.

\bibitem{standley2020tasks}
Trevor Standley, Amir Zamir, Dawn Chen, Leonidas Guibas, Jitendra Malik, and
  Silvio Savarese.
\newblock Which tasks should be learned together in multi-task learning?
\newblock In {\em International Conference on Machine Learning}, pages
  9120--9132. PMLR, 2020.

\bibitem{thrun1996discovering}
Sebastian Thrun and Joseph O'Sullivan.
\newblock Discovering structure in multiple learning tasks: The tc algorithm.
\newblock In {\em ICML}, volume~96, pages 489--497, 1996.

\bibitem{vandenhende2021multi}
Simon Vandenhende, Stamatios Georgoulis, Wouter Van~Gansbeke, Marc Proesmans,
  Dengxin Dai, and Luc Van~Gool.
\newblock Multi-task learning for dense prediction tasks: A survey.
\newblock {\em IEEE Transactions on Pattern Analysis and Machine Intelligence},
  2021.

\bibitem{xin2022current}
Derrick Xin, Behrooz Ghorbani, Justin Gilmer, Ankush Garg, and Orhan Firat.
\newblock Do current multi-task optimization methods in deep learning even
  help?
\newblock {\em Advances in Neural Information Processing Systems},
  35:13597--13609, 2022.

\bibitem{yang2020multi}
Ruihan Yang, Huazhe Xu, Yi~Wu, and Xiaolong Wang.
\newblock Multi-task reinforcement learning with soft modularization.
\newblock {\em arXiv preprint arXiv:2003.13661}, 2020.

\bibitem{yu2020gradient}
Tianhe Yu, Saurabh Kumar, Abhishek Gupta, Sergey Levine, Karol Hausman, and
  Chelsea Finn.
\newblock Gradient surgery for multi-task learning.
\newblock {\em arXiv preprint arXiv:2001.06782}, 2020.

\bibitem{yu2020meta}
Tianhe Yu, Deirdre Quillen, Zhanpeng He, Ryan Julian, Karol Hausman, Chelsea
  Finn, and Sergey Levine.
\newblock Meta-world: A benchmark and evaluation for multi-task and meta
  reinforcement learning.
\newblock In {\em Conference on Robot Learning}, pages 1094--1100. PMLR, 2020.

\bibitem{zamir2018taskonomy}
Amir~R Zamir, Alexander Sax, William Shen, Leonidas~J Guibas, Jitendra Malik,
  and Silvio Savarese.
\newblock Taskonomy: Disentangling task transfer learning.
\newblock In {\em Proceedings of the IEEE conference on computer vision and
  pattern recognition}, pages 3712--3722, 2018.

\bibitem{zhou2022convergence}
Shiji Zhou, Wenpeng Zhang, Jiyan Jiang, Wenliang Zhong, Jinjie Gu, and Wenwu
  Zhu.
\newblock On the convergence of stochastic multi-objective gradient
  manipulation and beyond.
\newblock {\em Advances in Neural Information Processing Systems},
  35:38103--38115, 2022.

\bibitem{zhugradient}
Shijie Zhu, Hui Zhao, Pengjie Wang, Hongbo Deng, Jian Xu, and Bo~Zheng.
\newblock Gradient deconfliction via orthogonal projections onto subspaces for
  multi-task learning.

\end{thebibliography}
\bibliographystyle{plain}
\clearpage
\appendix
\section{Gradient Manipulation Methods}
\label{sec::apx-grad-manip}
In this section, we provide a brief overview of representative gradient manipulation methods in multitask/multiobjective optimization. Specifically, we will also discuss the connections among these methods.
\paragraph{Multiple Gradient Descent Algorithm (\mgda{})~\citep{desideri2012multiple,sener2018multi}} The MGDA algorithm is one of the earliest gradient manipulation methods for multitask learning. In MGDA, the per step update $d_t$ is found by solving
        $$
        \max_{d \in \RR^m} \min_{i \in [k]} \nabla \ell_{i,t}^\top d - \frac{1}{2}\norm{d}^2.
        $$
As a result, the solution $d^*$ of \mgda{} optimizes the ``worst improvement" across all tasks or equivalently seeks an \emph{equal} descent across all task losses as much as possible. But in practice, \mgda{} suffers from slow convergence since the update $d^*$ can be very small. For instance, if one task has a very small loss scale, the progress of all other tasks will be bounded by the progress on this task. Note that the original objective in \eqref{eq:dual} is similar to the \mgda{} objective in the sense that we can view optimizing \eqref{eq:dual} as optimizing the log of the task losses. Hence, when we compare \famo{} against \mgda{}, one can regard \famo{} as balancing the \emph{rate} of loss improvement while \mgda{} balances the absolute improvement across task losses.

\paragraph{Projecting Gradient Descent (\pcgrad{})~\cite{yu2020gradient}} \pcgrad{} initializes $v^i_\text{PC} = \nabla \ell_{i,t}$, then for each task $i$, \pcgrad{} loops over all task $j \neq i$ (in a random order, which is crucial as mentioned in \citep{yu2020gradient}) and removes the ``conflict"
$$
v^i_{\text{PC}} \leftarrow v^i_\text{PC} - \frac{{v^i_{\text{PC}}}^\top \nabla \ell_{j,t}}{\norm{\ell_{j,t}}^2} \nabla \ell_{j,t}~~~~\text{if}~~~~ {v^i_{\text{PC}}}^\top \nabla \ell_{j,t} < 0.
$$
In the end, \pcgrad{} produces $d_t = \frac{1}{k} \sum_{i=1}^k v^i_\text{PC}$. Due to the construction, \pcgrad{} will also help improve the ``worst improvement" across all tasks since the ``conflicts" have been removed. However, due to the stochastic iterative procedural of this algorithm, it is hard to understand \pcgrad{} from a first principle approach.

\paragraph{Conflict-averse Gradient Descent (\cagrad{})~\cite{liu2021conflict}} $d_t$ is found by solving
        $$
        \max_{d \in \RR^m} \min_{i \in [k]} \nabla \ell_{i,t}^\top d ~~~~\text{s.t.}~~~~\norm{d - \nabla \ell_{0,t}} \leq c \norm{\nabla \ell_{0,t}}.
        $$
Here, $\ell_{0,t} = \frac{1}{k} \sum_{i=1}^k \ell_{i,t}$. \cagrad{} seeks an update $d_t$ that optimizes the ``worst improvement" as much as possible, conditioned on that the update still decreases the average loss. By controlling the hyperparameter $c$, \cagrad{} can recover \mgda{} ($c \rightarrow \infty$) and the vanilla averaged gradient descent ($c \rightarrow 0$). Due to the extra constraint, \cagrad{} provably converges to the stationary points of $\ell_0$ when $0 \leq c < 1$.

\paragraph{Impartial Multi-Task Learning (\imtlg{})~\cite{liu2020towards}} \imtlg{} finds $d_t$ such that it shares the same cosine similarity with any task gradients:
$$
    \forall i \neq j, ~~~ d_t ^\top \frac{\nabla \ell_{i,t}}{\norm{\nabla \ell_{i,t}}} = d_t^\top \frac{\nabla \ell_{j,t}}{\norm{\nabla \ell_{j,t}}},~~~\text{and}~~~d_t = \sum_{i=1}^k w_{i,t} \nabla \ell_{i,t},~~\text{for some}
~~{w}_t \in \S_k.$$
The constraint that $d_t = \sum_{i=1}^k w_{i,t} \nabla \ell_{i,t}$ is for preventing the problem from being under-determined. From the above equation, we can see that \imtlg{} ignores the ``size" of each task gradient and only cares about the ``direction". As a result, \textbf{one can think of \imtlg{} as a variant of \mgda{} that applies to the normalized gradients}. By doing so, \imtlg{} does not suffer from the straggler effect due to slow objectives. Furthermore, one can \textbf{view \imtlg{} as the equal angle descent}, which is also proposed in Katrutsa et al. \citep{katrutsa2020follow}, where the objective is to find $d$ such that
$$
\forall i \neq j, \qquad \cos(d, \nabla \ell_{i,t}) = \cos(d, \nabla \ell_{j,t}).
$$

\paragraph{\nashmtl{}\cite{navon2022multi}} \nashmtl{} finds $d_t$ by solving a bargaining game treating the local improvement of each task loss as the utility for each task:
$$
\max_{d \in \RR^m, \norm{d} \leq 1} \sum_{i=1}^k \log \big(\nabla \ell_{i,t}^\top d\big).
$$
Note that the objective of \nashmtl{} implicitly assumes that there exists $d$ such that $\forall~i,~~\nabla \ell_{i,t}^\top d > 0$ (otherwise we reach the Pareto front). It is easy to see that 
$$
\max_{\norm{d} \leq 1} \sum_{i=1}^k \log \big(\nabla \ell_{i,t}^\top d\big) = \max_{\norm{d} \leq 1} \sum_{i=1}^k \log \langle \frac{\nabla \ell_{i,t}}{\norm{\nabla \ell_{i,t}}}, d\rangle = \max_{\norm{d} \leq 1} \sum_{i=1}^k \log \cos\big(\nabla \ell_{i,t}, d\big).
$$
Therefore, due to the $\log$, \nashmtl{} also ignores the ``size" of task gradients and only cares about their ``directions". Moreover, denote $u_i = \frac{\nabla \ell_{i,t}}{\norm{\nabla \ell_{i,t}}}$. Then, according to the KKT condition, we know:
$$
\sum_i \frac{u_i}{u_i^\top d} - \alpha d = 0,\quad \alpha \geq 0 \qquad \Longrightarrow \qquad d = \frac{1}{\alpha} \sum_i \frac{1}{u_i^\top d} u_i.
$$
Consider when $k=2$, if we take the \emph{equal angle descent} direction: $d_\angle = (u_1 + u_2) / 2$ (note that as $u_1$ and $u_2$ are normalized, their bisector is just their average). Then it is easy to check that 
$$
d_\angle = \frac{1}{\alpha}\bigg( \frac{2}{u_1^\top (u_1 + u_2)} u_1 + \frac{2}{u_2^\top (u_1 + u_2)} u_2\bigg),~~\text{where}~~ \alpha = \frac{u_1^\top(u_1 + u_2)}{4} = \frac{u_2^\top(u_1 + u_2)}{4}.
$$
As a result, we can see that \textbf{when $k=2$, \nashmtl{} is equivalent to \imtlg{} (or the equal angle descent)}. However, when $k > 2$, this is not in general true.

\textbf{Remark} Note that all of these gradient manipulation methods require computing and storing $K$ task gradients before applying $f$ to compute $d_t$, which often involves solving an additional optimization problem. Hence, these methods can be slow for large $K$ and large model sizes.

\section{Amortizing other Gradient Manipulation Methods}
\label{sec::apx-fast-approx}
Although \famo{} uses iterative update on ${w}$, it is not immediately clear whether we can apply the same amortization easily on other existing gradient manipulation methods. In this section, we discuss such possibilities and point out the challenges.

\paragraph{Amortizing \mgda{}} This is almost the same as in \famo{}, except that \mgda{} acts on the original task losses while \famo{} acts on the log of task losses. 

\paragraph{Amortizing \pcgrad{}} For \pcgrad{}, finding the final update vector requires iteratively projecting one task gradient to the other, so there is no straightforward way of bypassing the computation of task gradients. 

\paragraph{Amortizing \imtlg{}} The task weighting in \imtlg{} is computed by a series of matrix-matrix and matrix-vector products using task gradients~\cite{liu2020towards}. Hence, it is also hard to amortize its computation over time.

Therefore, we focus on deriving the amortization for \cagrad{} and \nashmtl{}.

\paragraph{Amortizing \cagrad{}} For \cagrad{}, the dual objective is
\begin{equation}
\min_{{w} \in \S_k} F({w}) = g_{{w}}^\top g_0 + c \norm{g_{{w}}}\norm{g_0},
\label{eq:cagrad-fast-approx}
\end{equation}
where $g_0 = \nabla \ell_{0,t}$ and $g_{{w}} = \sum_{i=1}^k w_i \nabla \ell_i$. Denote 
$$
{G} = \begin{bmatrix}
        \nabla  \ell_{1,t}^\top \\
        \vdots \\
        \nabla  \ell_{k,t}^\top
    \end{bmatrix}.
$$
Now, if we take the gradient with respect to ${w}$ in \eqref{eq:cagrad-fast-approx}, we have:
\begin{equation}
    \pdv{F}{{w}} = {G}^\top g_0 + c\frac{\norm{g_0}}{\norm{g_{{w}}}}{G}^\top g_{{w}}.
\end{equation}
As a result, in order to approximate this gradient, one can separately estimate:
\begin{equation}
\begin{split}
{G}^\top g_0 &\approx \frac{{\ell}(\x) - {\ell}(\x - \alpha g_0)}{\alpha} \\
{G}^\top g_{{w}} &\approx \frac{{\ell}(\x) - {\ell}(\x - \alpha g_{{w}})}{\alpha} \\
\norm{g_0} &\approx \sqrt{ {1}^\top {G}^\top g_0 } \\
\norm{g_{{w}}} &\approx \sqrt{ {w}^\top {G}^\top g_{{w}} } \\
\end{split}.
\end{equation}
Once all these are estimated, one can combine them together to perform a single update on ${w}$. But note that this will require 3 forward and backward passes through the model, making it harder to implement in practice.

\paragraph{Amortizing \nashmtl{}} Per derivation from \nashmtl{}~\cite{navon2022multi}, the objective is to solve for ${w}$:
\begin{equation}
{G}^\top {G} {w} = {1} \oslash {w}. 
\end{equation}
One can therefore form an objective:
\begin{equation}
\min_{{w}} F({w}) = \norm{{G}^\top {G} {w} - {1} \oslash {w} }^2_2.
\end{equation}
Taking the derivative of $F$ with respect to ${w}$, we have
\begin{equation}
    \pdv{F}{{w}} = 2{G}^\top {G} \bigg({G}^\top g_{{w}} - {1} \oslash {w}\bigg) + 2 \bigg({G}^\top g_{{w}} - {1} \oslash {w}\bigg) \oslash ({w} \odot {w}).
\end{equation}
Therefore, to approximate the gradient of ${w}$, one needs to first estimate
\begin{equation}
{G}^\top g_{{w}} \approx \frac{{L}(\theta) - {L}(\theta - \alpha g_{{w}})}{\alpha} = {\eta}.
\end{equation}
Then we estimate
\begin{equation}
    {G}^\top {G} ({\eta} - {1} \oslash {w}) \approx \frac{{L}(\theta) - {L}(\theta - \alpha {G} ({\eta} - {1} \oslash {w}))}{\alpha}.
\end{equation}
Again, this results in 3 forward and backward passes through the model, let alone the overhead of resetting the model back to $\theta$ (requires a copy of the original weights).

In short, though it is possible to derive fast approximation algorithm to approximate the gradient update on ${w}$ for some of the existing gradient manipulation methods, it often involves much more complicated computation compared to that of \famo{}.

\section{\famo{} Pseudocode in PyTorch}
\label{sec::apx-pseudocode}
We provide the pseudocode for \famo{} in Algorithm~\ref{alg:pseudo}. To use \famo{}, one just first compute the task losses, call {\PyCode{get\_weighted\_loss}} to get the weighted loss, and do the normal backpropagation through the weighted loss. After that,  one call {\PyCode{update}} to update the task weighting.

\begin{algorithm}[t!]
\begin{flushleft}
\PyCode{class FAMO:}\\
\PyCode{def \_\_init\_\_(self, num\_tasks, min\_losses, $\alpha$=0.025, $\gamma$=0.001):}\\
\qquad\PyComment{min\_losses~~(num\_tasks,) the loss lower bound for each task.} \\
\PyCode{~~~~self.min\_losses = min\_losses}\\
\PyCode{~~~~self.xi = torch.tensor([0.0] * num\_tasks, requires\_grad=True)}\\
\PyCode{~~~~self.xi\_opt = torch.optim.Adam([self.xi], lr=$\alpha$, weight\_decay=$\gamma$)}\\
\PyCode{}\\
\PyCode{def get\_weighted\_loss(self, losses):}\\
\qquad\PyComment{losses~~(num\_tasks,)} \\
\PyCode{~~~~z = F.softmax(self.xi, -1)}\\
\PyCode{~~~~D = losses - self.min\_losses + 1e-8}\\
\PyCode{~~~~c = 1 / (z / D).sum().detach()}\\
\PyCode{~~~~loss = (c * D.log() * z).sum()}\\
\PyCode{~~~~return loss}\\
\PyCode{}\\
\PyCode{def update(self, prev\_losses, curr\_losses):}\\
\qquad\PyComment{prev\_losses~~(num\_tasks,)} \\
\qquad\PyComment{curr\_losses~~(num\_tasks,)} \\
\PyCode{~~~~delta = (prev\_losses - self.min\_losses + 1e-8).log() -}\\
\PyCode{~~~~~~~~~~~~(curr\_losses - self.min\_losses + 1e-8).log()}\\
\PyCode{~~~~with torch.enable\_grad():}\\
\PyCode{~~~~~~~~d = torch.autograd.grad(F.softmax(self.xi, -1),}\\
\PyCode{~~~~~~~~~~~~~~~~~~~~~~~~~~~~~~~~self.xi,}\\
\PyCode{~~~~~~~~~~~~~~~~~~~~~~~~~~~~~~~~grad\_outputs=delta.detach())[0]}\\
\PyCode{~~~~self.xi\_opt.zero\_grad()}\\
\PyCode{~~~~self.xi.grad = d}\\
\PyCode{~~~~self.xi\_opt.step}\\
\end{flushleft}
\caption{Implementation of \famo{} in PyTorch-like Pseudocode}
\label{alg:pseudo}
\end{algorithm}

\section{Toy Example}
\label{sec::apx-toy}
We provide the task objectives for the toy example in the following. The model parameter $\theta=(\theta_1, \theta_2) \in \RR^2$ and the task objectives are $L^1$ and $L^2$:
\begin{flalign*}
    L^1(\theta) &= 0.1\cdot (c_1(\theta)f_1(\theta) + c_2(\theta)g_1(\theta))~~\text{and}~~L^2(\theta) = c_1(\theta)f_2(\theta) + c_2(\theta)g_2(\theta),~\text{where}\\
    f_1(\theta) &= \log{\big(\max(|0.5(-\theta_1-7)-\tanh{(-\theta_2)}|,~~0.000005)\big)} + 6, \\
    f_2(\theta) &= \log{\big(\max(|0.5(-\theta_1+3)-\tanh{(-\theta_2)}+2|,~~0.000005)\big)} + 6, \\
    g_1(\theta) &= \big((-\theta_1+7)^2 + 0.1*(-\theta_2-8)^2\big)/10-20, \\
    g_2(\theta) &= \big((-\theta_1-7)^2 + 0.1*(-\theta_2-8)^2)\big/10-20, \\
    c_1(\theta) &= \max(\tanh{(0.5*\theta_2)},~0)~~\text{and}~~c_2(\theta) = \max(\tanh{(-0.5*\theta_2)},~0).
\end{flalign*}

\section{Experimental Results with Error Bars}
\label{sec::apx-result-error}
We followed the exact experimental setup from \nashmtl{}~\cite{navon2022multi}. Therefore, the numbers for baseline methods are taken from their original paper. In the following, we provide \famo{}'s result with error bars.
\clearpage
\begin{table*}[h!]
    \centering
    \resizebox{\textwidth}{!}{%
    \begin{tabular}{lrrrrrrrrrr}
    \toprule
      &  \multicolumn{2}{c}{Segmentation} & \multicolumn{2}{c}{Depth} & \multicolumn{5}{c}{Surface Normal} &\\
    \cmidrule(lr){2-3}\cmidrule(lr){4-5}\cmidrule(lr){6-10}
    \textbf{Method} &  \multirow{2}{*}{mIoU $\uparrow$} & \multirow{2}{*}{Pix Acc $\uparrow$} & \multirow{2}{*}{Abs Err $\downarrow$} & \multirow{2}{*}{Rel Err $\downarrow$} & \multicolumn{2}{c}{Angle Dist $\downarrow$} & \multicolumn{3}{c}{Within $t^\circ$ $\uparrow$}  & \dm{} $\downarrow$ \\
    \cmidrule(lr){6-7}\cmidrule(lr){8-10}
    & & & & & Mean & Median & 11.25 & 22.5 & 30  &\\
     \midrule
    \famo{} (mean)  & 38.88 & 64.90 & 0.5474 & 0.2194 & 25.06 & 19.57 & 29.21 & 56.61 & 68.98   & -4.10 \\
    \famo{} (stderr)& \fs{0.54} & \fs{0.21} & \fs{0.0016} & \fs{0.0026} &  \fs{0.06} & \fs{0.09} &  \fs{0.17} &   \fs{0.19} & \fs{0.14} & \fs{0.39} \\
    \bottomrule  
    \end{tabular}
    }
    \vspace{5pt}
    \caption{Results on NYU-v2 dataset (3 tasks). Each experiment is repeated over 3 random seeds and the mean is reported. The best average result is marked in bold. \mr{} and \dm{} are the main metrics for MTL performance.}
    \label{tab:apx-nyu-v2}
\end{table*}

\begin{table*}[h!]
    \centering
    \resizebox{\textwidth}{!}{%
    \begin{tabular}{lrrrrrrrrrrrr}
    \toprule
    \multirow{2}{*}{\textbf{Method}} & $\mu$ & $\alpha$ & $\epsilon_\text{HOMO}$ & $\epsilon_\text{LUMO}$ & $\langle R^2\rangle$ & ZPVE & $U_0$ & $U$ & $H$ & $G$ & $c_v$  & \multirow{2}{*}{ \dm{} $\downarrow$}\\
      \cmidrule(lr){2-12}
      & \multicolumn{11}{c}{MAE $\downarrow$} & \\
    \midrule 
    \famo{} (mean)  & 0.15 & 0.30 & 94.0 & 95.2 &  1.63 & 4.95 & 70.82 & 71.2 & 71.2 & 70.3 & 0.10 & 58.5 \\
    \famo{} (stderr)  & \fs{0.0046} & \fs{0.0070} & \fs{3.074} & \fs{2.413} &  \fs{0.0211} & \fs{0.0871} & \fs{2.17} & \fs{2.19} & \fs{2.19} & \fs{2.21} & \fs{0.0026} & \fs{3.26} \\
    \bottomrule 
    \end{tabular}
    }
    \vspace{5pt}
    \caption{Results on QM-9 dataset (11 tasks). Each experiment is repeated over 3 random seeds and the mean is reported. The best average result is marked in bold. \mr{} and \dm{} are the main metrics for MTL performance.}
    \label{tab:apx-qm9}
\end{table*}

\begin{table*}[h!]
    \centering
    \resizebox{0.7\textwidth}{!}{%
    \begin{tabular}{lrrrrrr}
    \toprule
   \multirow{4}{*}{\textbf{Method}} & \multicolumn{5}{c}{\textbf{CityScapes}} & \textbf{CelebA} \\
      \cmidrule(lr){2-6}\cmidrule(lr){7-7}
      &  \multicolumn{2}{c}{Segmentation} & \multicolumn{2}{c}{Depth} & \multirow{2}{*}{ \dm{} $\downarrow$}  & \multirow{2}{*}{ \dm{} $\downarrow$}\\
    \cmidrule(lr){2-3}\cmidrule(lr){4-5}
     &  mIoU $\uparrow$ & Pix Acc $\uparrow$ & Abs Err $\downarrow$ & Rel Err $\downarrow$ &  &  \\
     \midrule
    \famo{} (mean)     & 74.54 & 93.29 & 0.0145 & 32.59 & 8.13 &  1.21 \\
    \famo{} (stderr)   & \fs{0.11} & \fs{0.04} & \fs{0.0009} & \fs{1.06} & \fs{1.98} &  \fs{0.24} \\
    \bottomrule 
    \end{tabular}
    }
    \vspace{5pt}
    \caption{Results on CityScapes (2 tasks) and CelebA (40 tasks) datasets. Each experiment is repeated over 3 random seeds and the mean is reported. The best average result is marked in bold. \mr{} and \dm{} are the main metrics for MTL performance.}
    \label{tab:apx-cityscapes-and-celeba}
\end{table*}

\end{document}